\documentclass{article}
\usepackage[dvipsnames]{xcolor}
\usepackage[round]{natbib}
\usepackage{PRIMEarxiv}

\usepackage[utf8]{inputenc} 
\usepackage[T1]{fontenc}    
\usepackage{hyperref}
\hypersetup{
  citebordercolor= {0 1 0},      
}

\usepackage{url}            
\usepackage{booktabs}       
\usepackage{amsfonts}       
\usepackage{nicefrac}       
\usepackage{microtype}      
\usepackage{lipsum}
\usepackage{fancyhdr}       
\usepackage{graphicx}       
\graphicspath{{media/}}     

\usepackage{amsthm}
\theoremstyle{plain}
\newtheorem{theorem}{Theorem}[section]

\theoremstyle{definition}

\newtheorem{assumption}[theorem]{Assumption}
\theoremstyle{remark}

\usepackage{doi}

\usepackage{algorithm}
\usepackage{algpseudocode}
\usepackage{amsmath}
\usepackage{amsfonts}
\usepackage[labelfont=bf]{caption}
\usepackage{enumitem}
\usepackage{subcaption}
\usepackage{tikz}
\usepackage{booktabs}

\usepackage{textgreek}

\setlist[itemize]{left=0.3cm}

\newcommand*\circled[1]{\tikz[baseline=(char.base)]{
            \node[shape=circle,draw,inner sep=0.75pt] (char) {#1};}}

\title{Navigating in High-Dimensional Search Space: A Hierarchical Bayesian Optimization Approach}

\author{Wenxuan Li$^*$, Taiyi Wang\thanks{Equal Contribution.} , Eiko Yoneki  \\
Department of Computer Science and Technology\\
University of Cambridge\\
\texttt{wl446@cantab.ac.uk} \\
\texttt{\{Taiyi.Wang, eiko.yoneki\}@cl.cam.ac.uk
}  \\
}

\begin{document}
\maketitle

\begin{abstract}
Optimizing black-box functions in high-dimensional search spaces has been known to be challenging for traditional Bayesian Optimization (BO). In this paper, we introduce HiBO, a novel hierarchical algorithm integrating global-level search space partitioning information into the acquisition strategy of a local BO-based optimizer. HiBO employs a search-tree-based global-level navigator to adaptively split the search space into partitions with different sampling potential. The local optimizer then utilizes this global-level information to guide its acquisition strategy towards most promising regions within the search space.  A comprehensive set of evaluations demonstrates that HiBO outperforms state-of-the-art methods in high-dimensional synthetic benchmarks and presents significant practical effectiveness in the real-world task of tuning configurations of database management systems (DBMSs). 
\end{abstract}


\section{Introduction}\label{sec:intro}
Bayesian Optimization (BO) is a powerful method for optimizing \textbf{\textit{black-box functions}} whose internal structure is largely unknown and requires iterative evaluations for reaching the optima with significant computational cost. It has seen broad adoption in real-world scenarios, including Neural Architecture Search \citep{kandasamy2018bo-nas, ru2020inter-bo}, hyperparameter tuning \citep{hvarfner2022pi-bo, zimmer2021auto-pytorch}, and automatic configuration of database management systems \citep{van2017otter, zhang2021restune, zhang2022onlinetune}. However, it is well known that BO struggles to scale in high-dimensional search spaces due to reasons including the exponentially increased search space and over-exploration  \citep{binois2022HDGPBO, malu2021brief-HDBO, rashidi2024CTS}. 

Even worse, optimization over high-dimensional search spaces in real-world systems is more challenging. The task of DBMS configuration tuning exemplifies this challenge, which involves a high-dimensional configuration knob space, complex interdependencies between knobs, and noisy performance observations \citep{zhao2023dbms-survey}. Existing approaches towards this problem either rely on extensive pre-collected random samples \citep{van2017otter} or metadata of prior workloads \citep{zhang2021restune}, or restrict evaluations to manually selected low-dimensional spaces \citep{zhang2022onlinetune, cereda2021cgptuner}. Moreover, methods aiming to scale BO to high dimensionalities (to be discussed below) often lack evaluation on real-world systems, limiting insights into their practical effectiveness.

Various methods mitigate the curse of dimensionality by restricting sampling to the most promising region \citep{eriksson2019turbo, daulton2022MORBO, wang2020lamcts, munos2011doo, bubeck2011hoo, kim2020voo}. However, such greedy confinement can lead to inefficient use of structural information (e.g., how partitions are formed) and inflexible sampling scope, resulting in suboptimal performance—particularly when initial budgets are small. Fully exploiting these structural insights necessitates a hierarchical approach that integrates partitioning modules, enabling a balanced exploration-exploitation trade-off within feasible computational overheads. At the same time, the granularity of the partition—whether coarser or finer—directly affects both the quality of the extracted information and the overhead of maintaining it. This interplay underscores the need for an \emph{adaptive} partitioning mechanism: coarse partitions may overlook fine-grained optima, whereas overly fine ones risk high costs and premature overfitting to limited data.

Motivated by the limitations of existing high-dimensional BO methods and the need for effective real-world system tuning, we have the following contributions: 

\textbf{Hierarchical Framework for Bayesian Optimization}. We propose \textbf{Hi}erarchical \textbf{B}ayesian \textbf{O}ptimization (HiBO), a hierarchical BO variant that is the first to integrate global-level space partitioning information into the local BO model’s acquisition strategy instead of rigidly confining the sampling scope. HiBO is shown to provide $\sim$28\% more performance improvements than only using the local optimizer in high-dimensional DBMS configuration tuning tasks with limited sample budget. 

\textbf{Search Space Partitioning with Adaptive Control}. We introduce the search-tree-based space partitioning for the global-level navigator, recursively splitting the search space into the partitions with different sampling potential.  Besides, an adaptive mechanism on controlling the maximum search tree depth is proposed for effective exploration-and-exploitation trade-offs and computational cost reduction.

\textbf{Practical Evaluation on Real-world Benchmark}. Besides experiments on synthetic and simulated ``real-world" benchmarks \citep{jones2008mopta, wang2017rover}, we demonstrate HiBO's effectiveness and practical value via evaluation on a real-world DBMS configuration tuning task, comprehensively considering performance improvement, tuning time cost and failure rate.

\section{Background}
\subsection{Preliminaries}

Bayesian Optimization (BO) iteratively refines a probabilistic surrogate of a black-box objective function and selects new samples via an acquisition function \citep{candelieri2021gentle-bo}. Formally, it aims to solve:
\begin{equation}
    x^* = \arg\max_{x \in X}\, f(x),
\end{equation}
where $f(x)$ is expensive to evaluate and reveals little explicit structure. BO typically starts with a few random points, fits a surrogate model to the observed data, and then chooses the next query by maximizing an acquisition function. As more data is gathered, the surrogate improves, guiding increasingly effective sampling decisions.

Though BO has been recognized as a powerful approach across various problem domains, vanilla BO is known to be inherently vulnerable to the "curse-of-dimensionality" \citep{binois2022HDGPBO} when the search space has a large number of dimensions. As dimensionality increases, the average squared distance between uniformly sampled points grows, making it harder for distance-based surrogate models to capture correlations between data points \citep{rashidi2024CTS}.  Additionally, the increased distance can result in greater uncertainty assigned to data points by the estimated posterior, increasing acquisition functions unreasonably and trapping BO algorithms in cycles of excessive exploration ("over-exploration"; \citet{siivola2018correcting}).

\begin{table}[t]
\vskip 0.15in
\begin{center}
\begin{small}

\bgroup
\def\arraystretch{1.16}%
\begin{tabular}{p{2.8cm}p{1.4cm}p{1.6cm}p{2.7cm}p{3.2cm}}
\toprule
& HOO	& LA-NAS	& LA-MCTS	& \textbf{HiBO}    \\ 
\midrule
\textbf{Partitioning Op.}	& K-nary & LR	&\textcolor{ForestGreen}{Clustering \& Classification}	&\textcolor{ForestGreen}{Clustering \& Classification} \\ 

\textbf{Depth Control}	&$\times$	&$\times$	&$\times$	&\textcolor{ForestGreen}{Adaptive Control} \\ 
\textbf{Restricted Sampling}     & Yes	& Yes	& Yes	& \textcolor{ForestGreen}{No}	\\ 

\textbf{Sampling Strategy} & Random	& Random/BO	& BO	&\textcolor{ForestGreen}{BO with augmented acqf.}\\ 
\bottomrule
\end{tabular}
\egroup

\end{small}
\end{center}
\caption{Comparison between optimization algorithms based on space partitioning, where \textit{LR} refers to Linear Regression and \textit{acqf.} refers to acquisition function. \textit{Partitioning Op.} is the operation for splitting the search space (see Section \ref{sec:space-part} for more details).}
\label{table:algo-comp}
\vspace{-10pt}
\end{table}

\subsection{Related Work}

Several paradigms of approaches have been developed to scale Bayesian Optimization (BO) to high-dimensional search spaces. Some methods leverage structural assumptions about the objective function, such as additive kernels \citep{duvenaud2011additiveGP, kandasamy2015high-add, wang2018ebo} or subspace embeddings \citep{wang2016rembo, letham2020re, nayebi2019hesbo, papenmeier2022baxus}. Alternatively, variable-selection-based strategies are employed to identify the most important dimensions \citep{li2017dropoutBO, spagnol19hsic, shen2023vsbo, song2022mctsvs}. However, these assumptions of the input space may not always hold in practice, or involve uncertain user-estimated parameters for these assumptions, such as the existence of the latent subspace and its dimensionality respectively. 

Another line of work focuses on optimizing the sampling scope without such assumptions, which can be further divided into 1) local modeling \citep{eriksson2019turbo, daulton2022MORBO} and 2) space partitioning \citep{munos2011doo, bubeck2011hoo, wang2021la-nas, wang2020lamcts}. TuRBO, a representative of local modeling, maintains trust regions around the best samples to mitigate over-exploration. While prior space-partitioning-based methods focus on splitting the search space into different partitions and limiting the sampling scope to the most promising one. A detailed comparison space-partitioning-based optimization is presented in Table \ref{table:algo-comp}, which includes HOO \citep{bubeck2011hoo}, LA-NAS \citep{wang2021la-nas}, LA-MCTS \citep{wang2020lamcts} and HiBO. HOO traditionally applies axis-aligned K-nary partitioning for splitting the space. As more recent work, LA-NAS and LA-MCTS apply MCTS-inspired ideas to partition the search space based on machine learning techniques for more flexible space bifurcation. In our work, HiBO extends these MCTS-based concepts in LA-NAS and LA-MCTS by formalizing the notion of sampling potential and employing a general search tree instead of MCTS. HiBO's partitioning search tree can be adaptively adjusted during the search to balance exploration-exploitation trade-off, and reduce unnecessary computational cost. Additionally, HiBO does not restrict sampling scope to the only specific region, allowing greater flexibility and more comprehensive consideration collected samples.

\section{Methodology: Hierarchical Bayesian Optimization}
\begin{figure*}[t]
    \centering
    \includegraphics[width=0.9\linewidth]{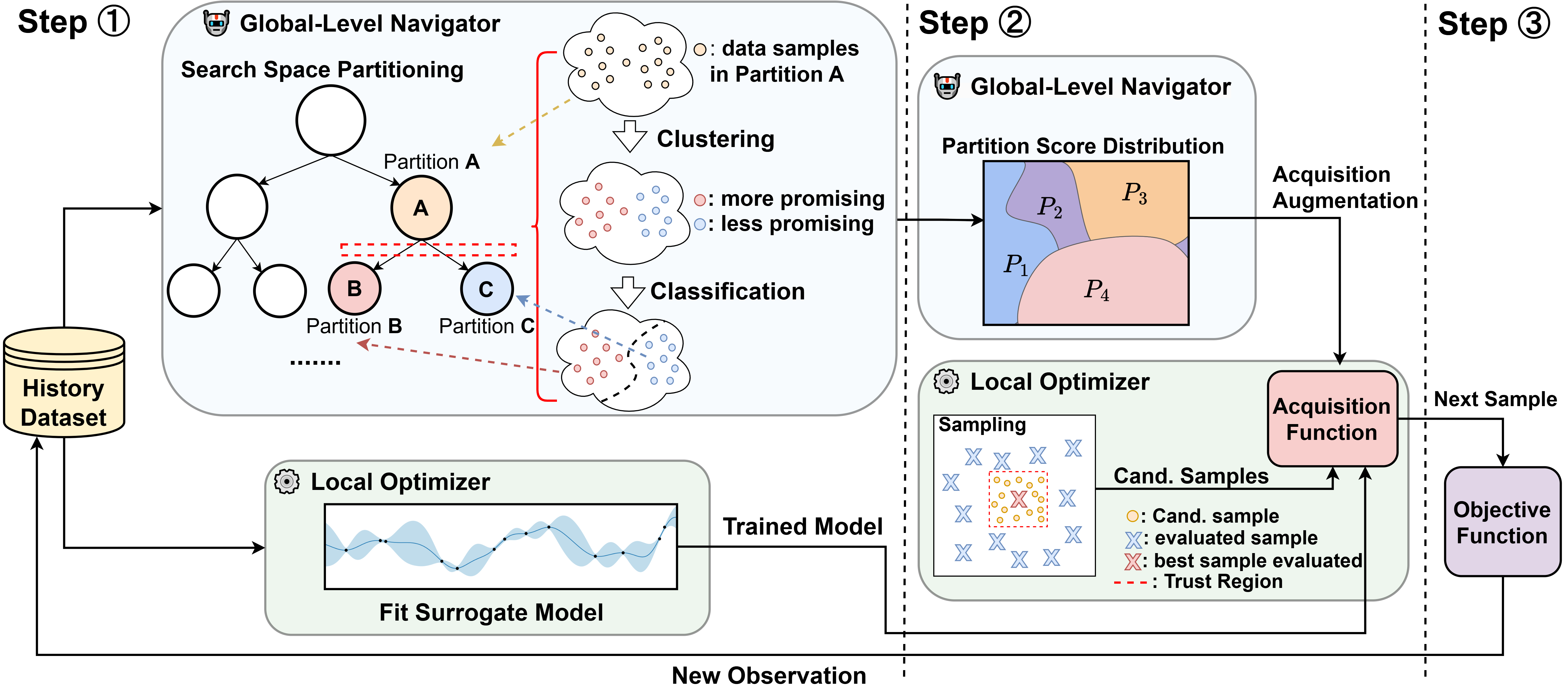}
    \caption{Illustration of the high-level workflow of HiBO within each optimization iteration.}
    \label{fig:HiBO-overview}
    \vskip -0.1in
\end{figure*}
\subsection{Overview}
In this section, we will introduce the high-level design of HiBO. HiBO is a \textbf{hierarchical} BO approach consisting of two parts: 1) a \textbf{global-level navigator}, which conducts search-tree-based search space partitioning based on collected data with dynamic and adaptive constraints posed when constructing the search tree; 2) a \textbf{local optimizer} with a BO-based algorithm as its core. Specifically we choose TuRBO (TuRBO-1; \citet{eriksson2019turbo}) as the local optimizer throughout this paper considering its practical effectiveness and simplicity for implementation. 

After the initial sampling, HiBO enters the main optimization loop. The high-level workflow for each iteration is outlined below, where Figure \ref{fig:HiBO-overview} illustrates this procedure: 
\begin{itemize}
    \item \textbf{Step \circled{1}}: Based on the history dataset, the global-level navigator constructs the search tree to partition the search space and generates a distribution of partition scores, while the local optimizer trains its surrogate model. The details of search space partitioning is explained in Section \ref{sec:space-part}.
    \item \textbf{Step \circled{2}}: The local optimizer randomly samples a set of candidate points within the trust region and calculates the modified acquisition function values based on the trained surrogate model and partition-based scores.
    \item \textbf{Step \circled{3}}: The candidate with the highest adapted acquisition function value is selected for objective function evaluation, and the new observation is added to the history dataset.
\end{itemize}

The key components of HiBO mentioned in the procedure above will be explained in the following subsections.

\subsection{Global-level Navigator: Adaptive Search Space Partitioning}\label{sec: global-navig}

The core motivation of introducing this global-level navigator is to partition the search space in a way maximizing the difference in sampling potential \footnote{Note that we intuitively describe the \textbf{\textit{sampling potential}} of a search space partition as the potential of sampling high-performing data points within this partition in the following sections.} across resulting partitions.  The navigator generates a distribution of sampling potential over these partitions, which will be utilized by the local optimizer for augmenting its acquisition strategy to boost its optimization efficiency in high-dimensional search spaces (see Section \ref{sec: weighted-acqf}). Besides the space partitioning itself, the design of the global-level navigator includes mechanisms for reducing computational cost and balancing trade-offs between exploration and exploitation.

\subsubsection{Data-Driven Search-Tree-based Space Partitioning}\label{sec:space-part}

During each iteration, the navigator constructs a search tree in a data-driven manner, where each node represents a partition of the original search space and is associated with the samples falling into this partition. Starting from the root node representing the entire search space, this method recursively splits the space into smaller and more specific partitions, as illustrated in \textbf{Step \circled{1}} of Figure \ref{fig:HiBO-overview}. 

To partition a node, clustering techniques like K-Means \citep{macqueen1967kmeans} are used to group the samples from the history dataset that fall within this node's partition into two clusters, assigning each sample a binary label. A classification model is then trained on sample features to predict the cluster labels and hence learns a decision boundary dividing the current search space partition into two parts. Each of these two parts thus forms a bifurcation of the original partition. Considering clustering is performed on both sample features and their performance data, samples in the two resulting sub-partitions are expected to differ significantly in terms of both performance and feature similarity, enabling the differentiation of valuable data points within different partitions.

By bifurcating the original partition into two sub-partitions, two child nodes are generated, each corresponding to one of the sub-partitions and containing the samples that fall within them.  Recursively repeating this ML-based bifurcation starting from the root node will finally result in a binary tree of nodes, where each leaf node corresponds to a sequence of split operations and represents a highly specific partition of the original search space. These refined partitions provide global-level insights into which regions of the search space are likely to be valuable for further sampling.

\subsubsection{Sampling Potential Evaluation via UCT}

Once the search tree is constructed, HiBO utilizes Upper Confidence bounds applied to Trees (UCT; \citet{kocsis2006UCT}) to evaluate each node's sampling potential by balancing exploitation and exploration:
\vskip -0.15in
\begin{equation} \label{eq:UCT}
    UCT_j = \hat{v_j} + 2C_p\sqrt{2\log n_p /n_j}
\end{equation}
Here, $C_p$ is a hyperparameter controlling the exploration-exploitation trade-off, $n_j$ and $n_p$ denote the number of visits to node $j$ and its parent node respectively, and $\hat{v}_j$ represents the average objective value within node $j$. By integrating UCT, HiBO prioritizes data points from partitions with higher average performance while still exploring less-visited regions of the search space, striking a balance between exploitation and exploration. 

In summary, the essence of HiBO's search-tree-based partitioning strategy involves: 1) utilizing machine learning techniques to partition the search space, aiming to maximize the difference of sampling potential across partitions; and 2) applying UCT to quantitatively assess such potential of each partition while balancing the exploration and exploitation.

\subsubsection{Dynamic and Adaptive Search Tree Construction}\label{sec:ssp-rules}
In addition to the procedures described for search space partitioning, we address two additional concerns regarding the proposed strategies:
\vskip -0.2in
\begin{itemize} 
    \setlength\itemsep{-0.18em}
    \item \textbf{Computational Cost}. Building a search tree at each iteration can be computationally expensive, where multiple classifiers such as Support Vector Machines \citep{boser1992svm} need to be trained and are used for inferring labels of thousands of candidate data points. 
    \item \textbf{Balance of Exploration and Exploitation} The depth of the tree, determined by the number of bifurcations from root to leaf, affects the range of the search space represented by each leaf node. As the tree deepens, leaf nodes represent smaller and more specific partitions, potentially leading to insufficient exploration.
\end{itemize}
The two concerns necessitate a dynamic and adaptive strategy to control the maximum depth of the constructed search tree during optimization. Towards this target, inspired by TuRBO's approach \citep{eriksson2019turbo} for exploration control via dynamic trust region adjustment, HiBO allows the tree to expand or shrink based on consecutive successes or failures. Consecutive successes reduce the number of bifurcation and hence increases the range covered by each leaf node, thereby promoting more exploration by considering a broader set of samples with equal partition-score weights.

\begin{figure}[t]
    \centering
    \begin{subfigure}[b]{0.5\textwidth}
        \centering
        \includegraphics[width=\linewidth]{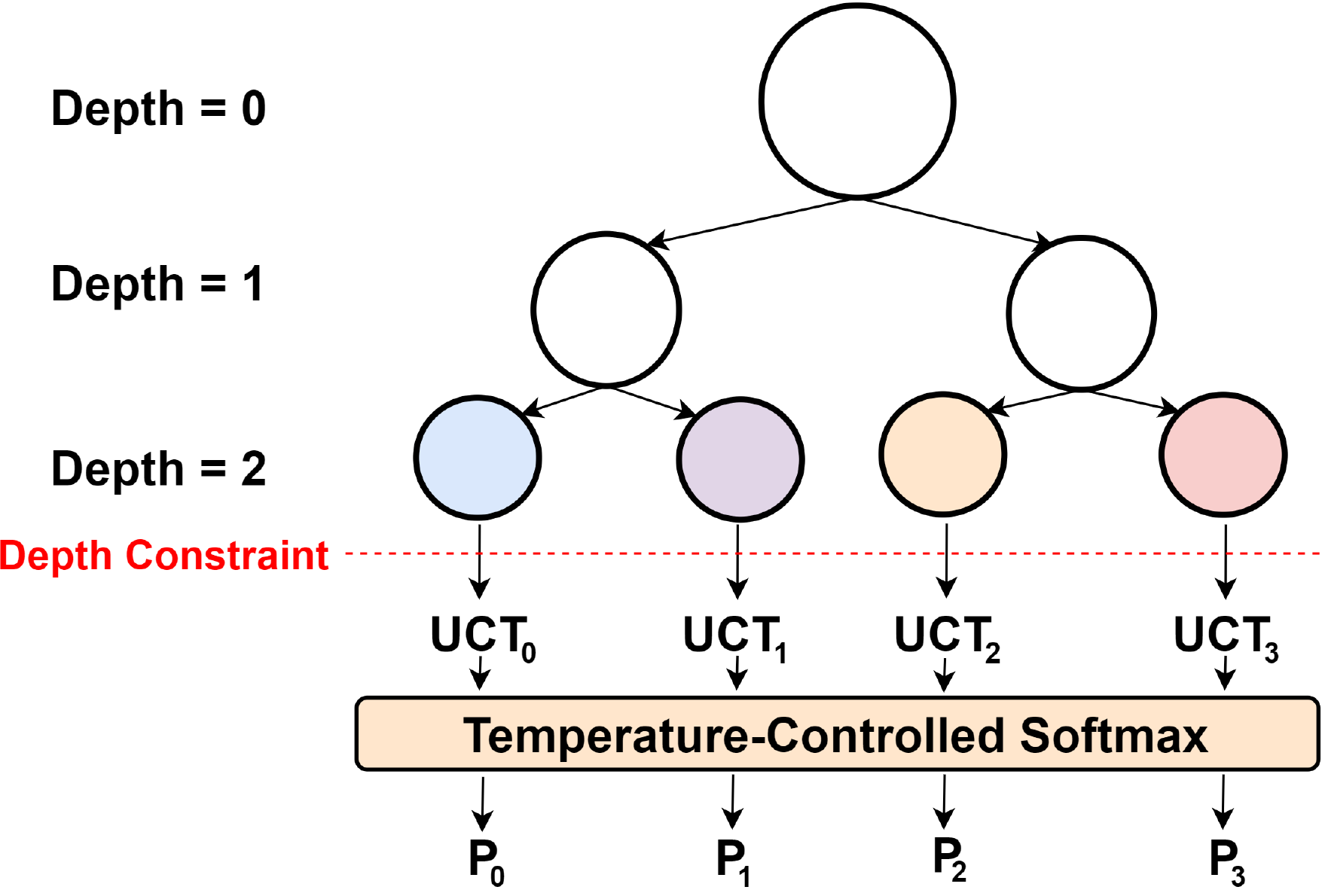}
        \caption{Calculation of partition scores from an example search tree.}
        \label{fig:temp-softmax}
    \end{subfigure}
    \hfill
    \begin{subfigure}[b]{0.45\textwidth}
        \centering
        \includegraphics[width=\linewidth]{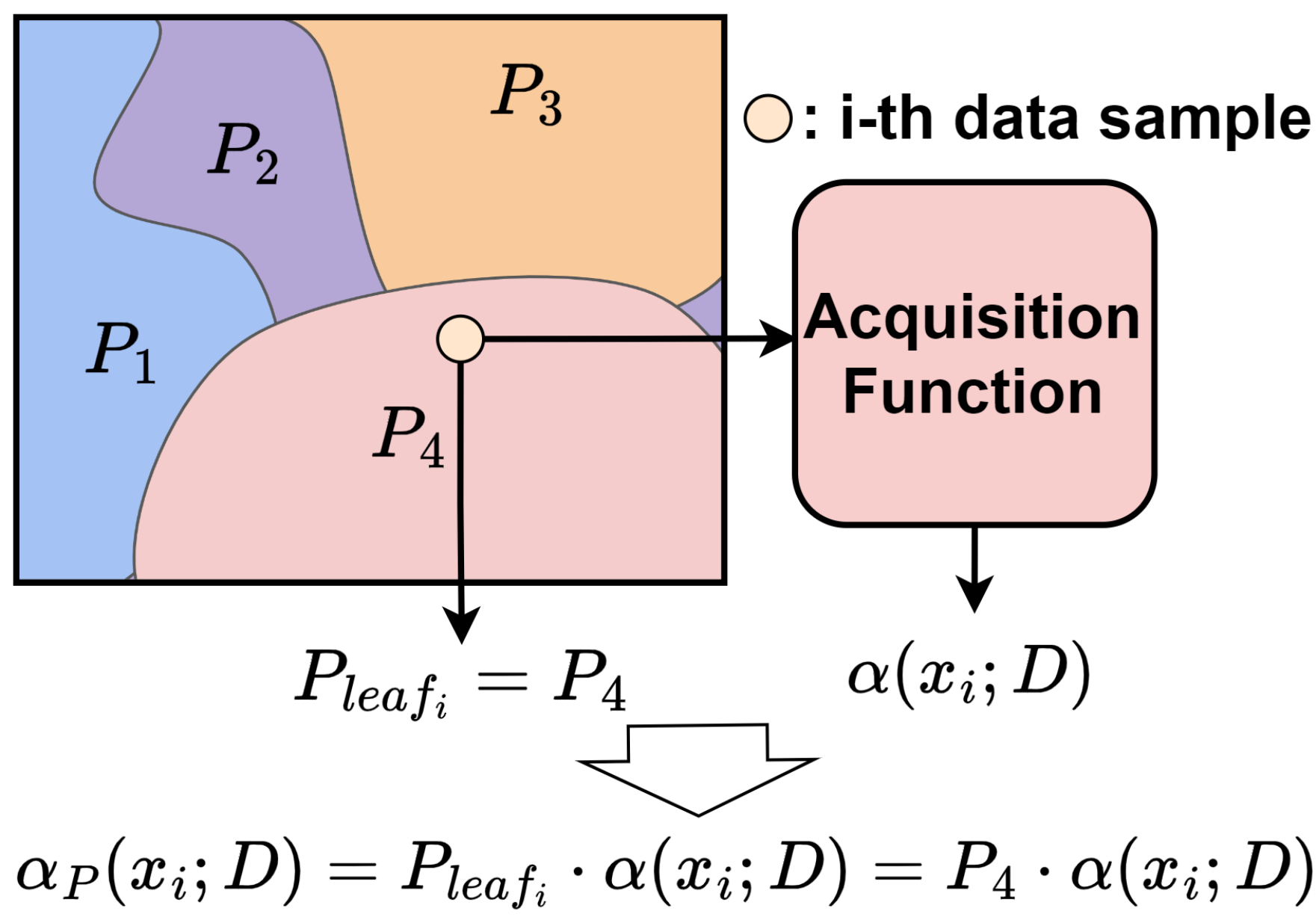}
        \caption{Calculation of the weighted acquisition function value corresponding to a sample point (see Section \ref{sec: weighted-acqf}).}
        \label{fig:weighted-acqf}
    \end{subfigure}
    \caption{Illustration of how the constructed search tree is integrated into the local optimizer's acquisition strategy.}
    \label{fig:hibo}
\end{figure}

The additional rules for building the tree are explained as follows:
\begin{itemize}
    \item \textbf{Rule 1} (Breadth-First Search Tree Construction)  Strictly follow the breadth-first order to build the tree layer by layer. Track the depth of each newly split child node and stop splitting once they reaches the current maximum tree depth, which is shown by the red line in Figure \ref{fig:temp-softmax};
    \item \textbf{Rule 2} (Adaptive Maximum Tree Depth) After a certain number of consecutive successes (improvements over the current best), reduce the maximum tree depth to broaden the range covered by each leaf node. Conversely, increase the maximum depth after consecutive failures for more focused biasing and hence more exploitation;
    \item \textbf{Rule 3} (Restart When Too Deep) If the maximum tree depth goes beyond a predefined threshold, instruct the local optimizer to restart with a new round of initial sampling and reset the maximum tree depth.
\end{itemize}

This dynamic and adaptive control on the maximum depth of the search tree aims to prevent unnecessary tree growth, reducing computational cost while maintaining a balance between exploration and exploitation. A further validation of the effectiveness of such adaptive tree depth settings are shown in Appendix \ref{app: ada-depth}.

\subsection{Local Optimizer: Partition-Score-Weighted Acquisition}\label{sec: weighted-acqf}
This section presents the design for the local optimizer to integrate the global space partitioning information into its acquisition strategy, with key operations illustrated in Figure \ref{fig:hibo}.

After the tree is built, all leaf nodes together represent the entire search space, with each of them serving as an information source of the global partitioning. Each leaf node $j$ is assigned with a UCT score $UCT_j$ following Equation \ref{eq:UCT}, reflecting the sampling potential of each partition of the whole search space. For effectively utilizing them, we calculate the \textit{partition score} $P_i$ for each node $\Omega_i$ by a temperature-controlled softmax function to scale all scores into a positive and reasonable range:
\begin{equation}\label{eq: softmax}
    P_{j} = \frac{exp(UCT_j / \tau)}{\sum_{j'}exp(UCT_{j'} / \tau)}
\end{equation}
where $\tau$ is a temperature parameter controlling the smoothness of the output distribution \citep{hinton2015distilling}. Lower $\tau$ values make the distribution sharper and strengthen the bias towards samples from partitions with greater UCT values.  

Given the partition score, the \textit{local optimizer} modifies its the acquisition function value $\alpha(x;D)$  for $i$-th candidate data point $x_i$, conditioned on the history dataset $D$, as shown in Figure \ref{fig:weighted-acqf}: 
\begin{equation*}
    \alpha_P(x_i; D) = \alpha (x_i; D) \cdot P_{leaf_{x_i}}
\end{equation*}
$leaf_{\bf{x}}$ denotes the leaf node containing $\bf{x}$. With the weighted acquisition function value, the next sample $x_{next}$ to be evaluated is the one maximizing: 
\begin{equation*}
    x_{next} = \underset{x \in X}{arg\: max} \; \alpha_P(x;D)
\end{equation*}
By weighing the acquisition function with the partition scores, the method prioritizes points from regions of the search space with higher sampling potential informed by the global-level navigator, while still considering the posterior estimated by the local surrogate model without constraining sampling to certain partitions. Such an augmentation of the original acquisition is aimed to make the search both globally informed and locally refined, and hence leads to a more effective exploration of the search space. 

\section{Theoretical Analysis}
\label{sec:theory-partition}


This section analyzes HiBO's theoretical properties, focusing on sample efficiency and scalability. We derive an upper bound for samples in the most promising partition and discuss computational complexity. More discussions on comparisons with prior works \citep{wang2021la-nas, wang2020lamcts} and scalability are provided in Appendix \ref{app:add_theory}.

\subsection{Bounding the Most Promising Partition}
\label{sec:bounding-most-promising}



At the root node of the partition tree (the global search domain $\Omega$), we define a \emph{partition error} $\delta$ to capture how far the split deviates from a perfect $N/2$--$N/2$ division when separating samples with higher and lower performance.

Formally, $\delta$ is the number of samples assigned to the subset with lower average performance that actually belong to the higher-performance set (or vice versa).   By Assumption \eqref{eq:mean-med-bound} in Appendix \ref{assump:density}, it can be proved that at most $\delta$ samples fall into such a misclassified subpartition, so that each partition is of size at most $\frac{N}{2} \;+\; \delta$. We now extend this reasoning to a \emph{tree of height~$h$}, constructed by \emph{recursively} subdividing the most promising region at each node based on the global-level navigator’s UCT-driven scores:

\begin{theorem}\label{thm:leaf-bound}
\textbf{(Bound on Most Promising Leaf).}
Consider a partition tree of height $h$ built by recursively splitting the sub-domain with the highest partition score.  Let $\delta_{\max}$ be the maximum partition error of any node along that ``leftmost'' path.  Then the sub-domain (leaf) at depth $h$ on that path contains at most
$
2\,\delta_{\max}\,\Bigl(1 - \frac{1}{2^h}\Bigr)\;+\;\frac{N}{2^h}
$
solutions.
\end{theorem}

\begin{proof}
Let $\delta_1, \delta_2, \ldots, \delta_h$ be the partition errors at successive tree levels along the path to the most promising leaf.  At the root, we split $\Omega$ into two sets of size at most $N/2 + \delta_1$ and $N/2 - \delta_1$, respectively.  By  Assumption \eqref{eq:mean-med-bound}, the left child takes the larger set, hence it has size $N/2 \;+\; \delta_1.$ At the next level, we apply the same argument to that child’s sub-domain of size $N/2 + \delta_1$, resulting in two further splits.  In the worst case, the sub-domain with the highest partition score again takes the larger portion: $
\frac{N}{2} \;+\; \delta_1 \;\;\longrightarrow\;\; \frac{N}{2^2} \;+\; \frac{\delta_1}{2} \;+\; \delta_2 
\quad\text{(at level 2)},$ and so on, possibly incurring a partition error $\delta_j$ at each stage.  Recursively unrolling this for $h$ splits, we get
\[
\underbrace{\delta_h \;+\; \frac{\delta_{h-1}}{2} \;+\; \frac{\delta_{h-2}}{2^2} \;+\;\dots \;+\; \frac{\delta_1}{2^{h-1}}}_{\text{accumulated partition error}}
\;+\;\frac{N}{2^h}.
\]
Defining $\delta_{\max} = \max\{\delta_1,\delta_2,\dots,\delta_h\}$ and bounding each $\delta_j$ by $\delta_{\max}$ yields:
\begin{align*}
\delta_h + \frac{\delta_{h-1}}{2} + \dots + \frac{\delta_1}{2^{h-1}}
&\;\;\le\;\;
\delta_{\max}\,\Bigl(1 \;\;+\dots\;\frac{1}{2^{\,h-1}}\Bigr) \\
&\;=\;
2\,\delta_{\max}\,\Bigl(1 - \frac{1}{2^h}\Bigr).
\end{align*}
Combining this with $\tfrac{N}{2^h}$ completes the proof,
\[
\delta_h + \dots + \frac{\delta_1}{2^{h-1}}
\;+\;\frac{N}{2^h}
\;\;\le\;\;
2\,\delta_{\max}\,\Bigl(1 - \frac{1}{2^h}\Bigr)\;+\;\frac{N}{2^h}.
\]
\vskip -0.2in
\end{proof}

The bound above shows that, as $h$ increases, the size of the subpartition containing the most promising solutions shrinks at a rate \(\tfrac{N}{2^h}\), up to an additive term that depends on the worst-case partition error $\delta_{\max}$.  Hence, in an idealized setting with small $\delta_{\max}$, HiBO zooms in on a small fraction of $\Omega$ at exponential speed in terms of $h$. The assumption can be found in Appendix \ref{assump:density}.

\subsection{Implications}
\textbf{Exponential Pruning of Sub-Optimal Regions.}
Theorem~\ref{thm:leaf-bound} shows HiBO can isolate high-performing samples at a rate of \(\tfrac{N}{2^h}\) with tree depth $h$. While computational cost grows exponentially with $h$, the quality of the best partition approaches the true optimum in $\Omega$, accelerating pruning compared to random sampling's $\frac{N}{2}$ complexity.

\textbf{Impact of Accurate Partitioning (\boldmath$\delta_{\max}$).}
The partition accuracy $\delta_{\max}$ improves as more observations refine the UCT scores. This explains HiBO's enhanced performance in larger search spaces: once reliable UCT estimates are established, the advantage over uniform exploration becomes more significant.

\section{Evaluation}
In this section, we present our evaluation experiments including: 1) synthetic benchmarks and 2) a real-world case study of automatic DBMS configuration tuning to comprehensive measure the effectiveness and practical value of HiBO on high-dimensional problems; 3) a comparative visual analysis of HiBO's acquisition function distribution across iterations to validate the behaviours of HiBO; and 4) empirical validation on the scalability of HiBO with respect to increased dimensionality of the search space.

\subsection{Synthetic Benchmarks}
\subsubsection{Experiment Setup}
To evaluate HiBO on synthetic benchmarks, we compare it with several related approaches, including vanilla GP-based BO, TuRBO \citep{eriksson2019turbo}, LA-MCTS \mbox{\citep{wang2020lamcts}}, MCTS-VS \citep{song2022mctsvs} and CMA-ES \citep{lozano2006cma-es}.  TuRBO is selected for validating the effects of the extra space partitioning information. We compare with LA-MCTS, the current SOTA in space-partitioning-based high-dimensional BO, to validate our design's efficiency, while excluding prior methods like HOO and LA-NAS for simplicity. We also include MCTS-VS, which uses MCTS for variable selection. Besides, for evaluating the effectiveness of HiBO on guiding other local optimizer, we introduce HiBO-GP, where the GP-based BO serves as the local optimizer of HiBO. 

\begin{figure}[t]
    \centering
    \includegraphics[width=0.9\linewidth]{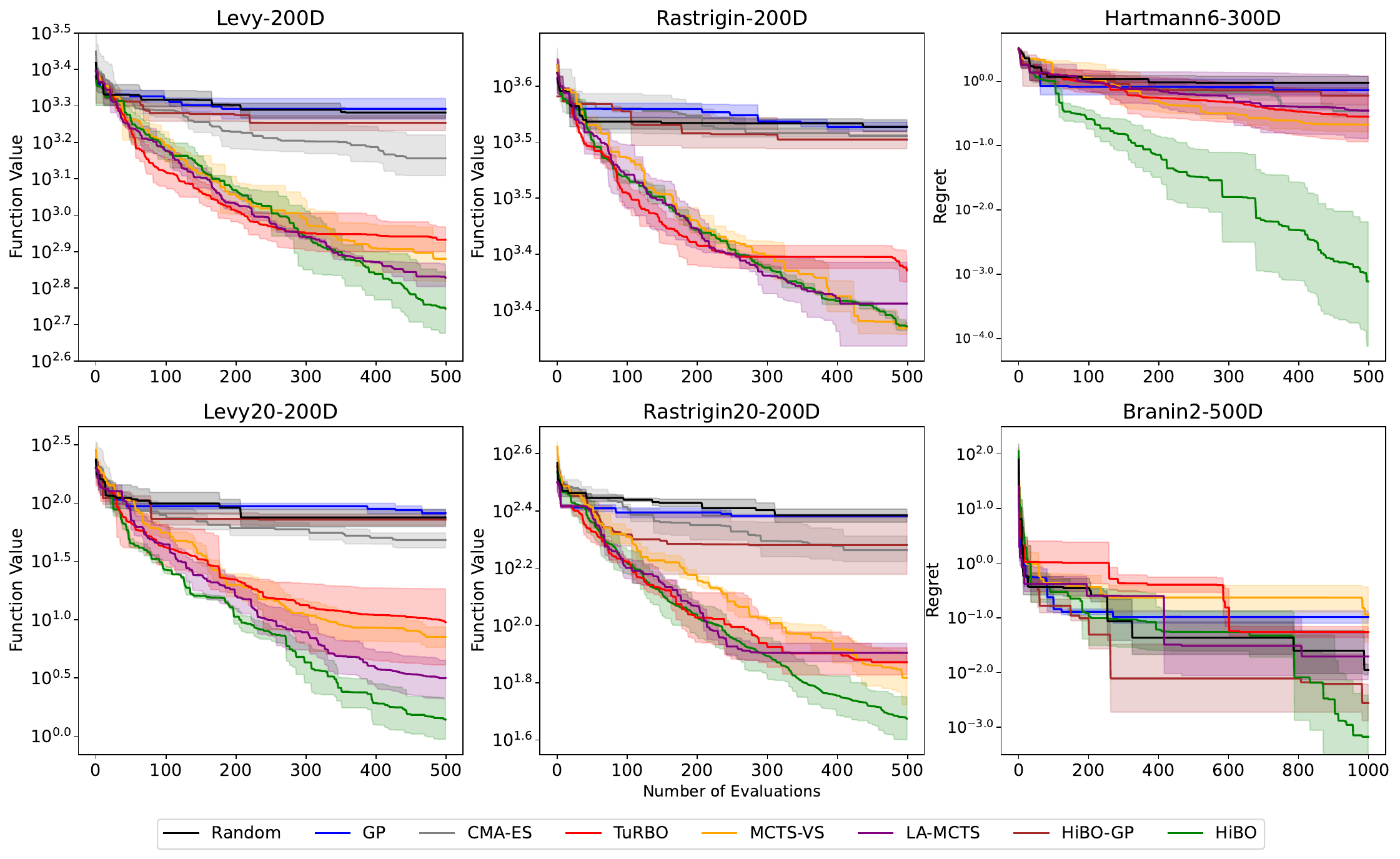}
    \caption{Evaluation results of algorithms on selected synthetic benchmarks.}
    \label{fig:synthetic}
\end{figure}

We evaluate these algorithms on six high-dimensional synthetic benchmarks. To evaluate how well the algorithms focus on relevant dimensions in sparse tasks, we evaluate Branin2-500D, Hartmann6-300D, Rastrigin20-200D, and Levy20-200D\footnote{'Benchmark(x)-(y)D' indicates $x$ effective dimensions out of $y$, with the remaining $(y-x)$ being dummy dimensions.}. Here \textbf{\textit{regret}} is measured for Hartmann6 and Branin2 benchmarks and defined as the difference between the optimal objective value and the algorithm's achieved value at a given iteration, with \textbf{lower regret indicating better algorithm performance.} We run 1000 iterations for Branin2-500D and 500 iterations for the others, starting with 40 and 20 iterations of random sampling, respectively. The additional configuration details for these algorithms are shown in Appendix \ref{app:syn-config}.


As observed from results on all these benchmarks (Figure \ref{fig:synthetic}), HiBO consistently achieves the best result across all benchmarks. Its superior performance compared to TuRBO demonstrates the advantage of incorporating global space partitioning information into the local model's acquisition strategy. HiBO also outperforms LA-MCTS on most benchmarks, indicating the effectiveness of navigating the local optimizer with partitioning information rather than limiting the sampling scope to the most promising partition. The performance gap with other algorithms widens on sparse benchmarks and becomes greatest on Hartmann6-300D, where HiBO achieves over two magnitudes lower regrets than others. This indicates HiBO's strong capacity to identify the most promising regions in high-dimensional tasks.

\subsection{Real-world Case Study: DBMS Configuration Tuning}
\subsubsection{Evaluation Background}\label{sec:dbms-back}

Besides evaluation on synthetic benchmarks, we evaluate the practical value of HiBO on a more complex real-world application, specifically automatic DBMS configuration tuning. The performance of DBMS highly depends on the settings of their configuration knobs, which control various aspects of their behaviours such as memory allocation, I/O behaviors, and query optimization. However, default configurations are typically suboptimal and require further tuning for optimal performance \citep{zhao2023dbms-survey}. We select DBMS configuration tuning especially for the following considerations: 1) \textbf{High-dimensionality}. DBMS configuration tuning involves a high-dimensional search space with over 100 configuration knobs; 2) \textbf{Knob Correlation}. Many of these configuration parameters are interdependent, where certain knobs are correlated and can lead to combined effects on the performance; 3) \textbf{Noisy Observations}. Measurements of performance data in real-world environments are much noiser compared to cases in synthetic benchmarks. These factors make DBMS configuration tuning an ideal target for evaluating practical values of algorithms in real-world applications. In comparison, though not limited to simpler synthetic functions, commonly-seen ``real-world" benchmarks in literature are still large based on simulation \citep{jones2008mopta, wang2017rover, eriksson2021saasbo}. For comprehensiveness, we still included the evaluation on Mopta08 \citep{jones2008mopta} and Rover Trajectory Planning \citep{wang2017rover} in Appendix \ref{app: eval_real}, including Mopta08 and Rover Trajectory Planning \citep{wang2017rover}.

\begin{figure}[t]
    \centering
    \includegraphics[width=0.88\linewidth]{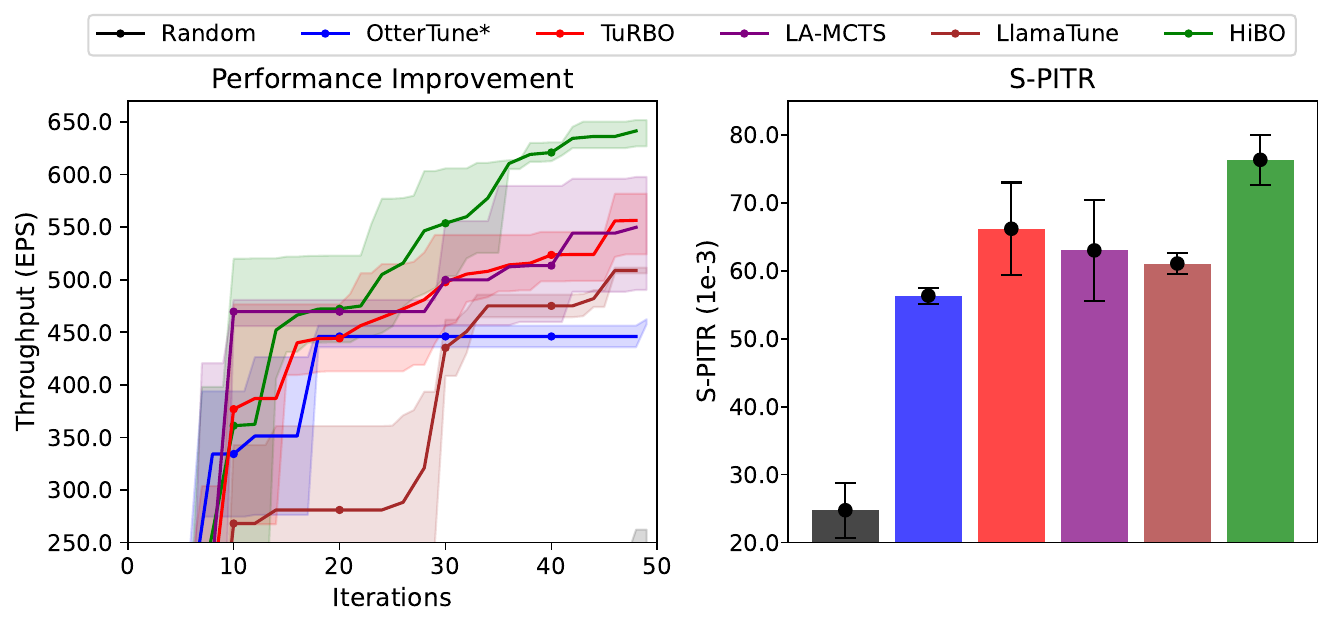}
    \caption{\textbf{Left}: Performance (throughput) evaluation of PostgreSQL being tuned by different algorithms on SysBench. Note that the figure for SysBench omits the low-performing part (throughput \textless \space 250 EPS) for readability. \textbf{Right}: S-PITR measurements of experiments done with the selected algorithms, which is explained with details in Section \ref{sec:db-res}.}
    \label{fig:dmbs-res}
\end{figure}

\subsubsection{Experiment Setup}\label{sec:dbms-setup}

Our evaluation focuses on the effectiveness of HiBO for DBMS configuration tuning without relying on extra sampling besides its own basic initial sampling. We choose PostgreSQL as the target DBMS considering its accessibility and general applicability. From \citet{PostgreSQLCO.NF}, we manually selected 110 knobs for tuning, excluding those related to debugging, network connections, and authorization. SysBench \citep{kopytov2004sysbench} is applied as a standard OLTP workload in the experiment with throughput (measured in events per second, EPS) as the target metric to be maximized. We used a limited sample budget of 50 iterations for simulating scenarios in production environments.

Selected algorithms for comparison include related BO methods and approaches specifically designed for DBMS configuration tuning: 1) Random search with uniform sampling; 2)  OtterTune \citep{van2017otter} without workload characterization and knob selection, noted as OtterTune$^*$; 3) TuRBO, 4) LA-MCTS, and 5) LlamaTune \citep{kanellis2022llamatune}. The details of algorithm selection, explanation and experiment configurations can be found in Appendix \ref{app:dbms-config}.

\begin{figure}[t]
    \centering
    \begin{subfigure}[b]{\textwidth}
        \centering
        \includegraphics[scale=0.3]{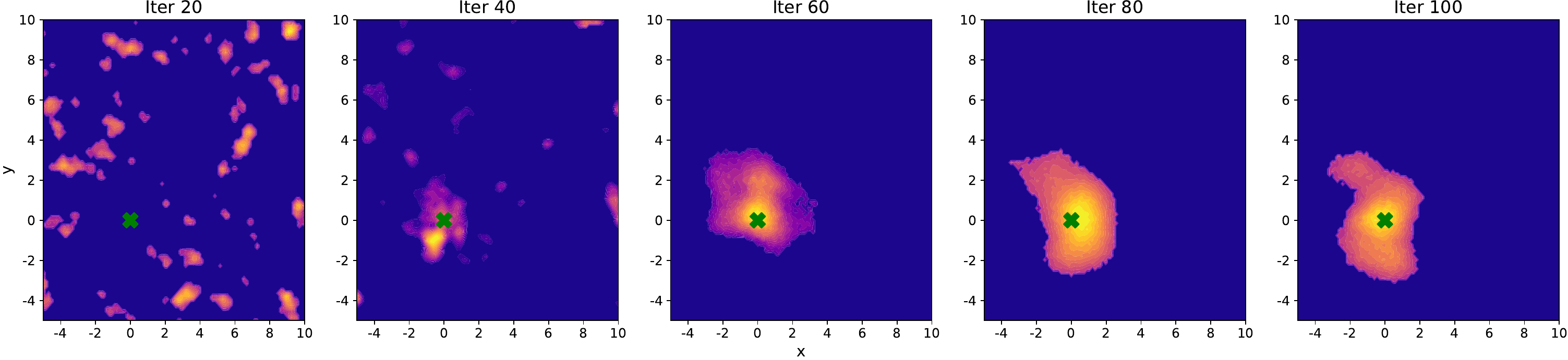}
        \caption{Distribution of partition-score-weighted acquisition scores (HiBO).}
        \label{fig:navig-dist}
    \end{subfigure}
    \vfill
    \begin{subfigure}[b]{\textwidth}
        \centering
        \includegraphics[scale=0.3]{figs/ackley2d_navig_dist.pdf}
        \caption{Distribution of vanilla acquisition scores (TuRBO).}
        \label{fig:vanilla-dist}
    \end{subfigure}
    \caption{Visualization of distribution of partition-score-weighted acquisition scores (HiBO) and vanilla acquisition scores (TuRBO) on the first two dimensions of Ackley2-200D across iterations. Lighter color indicates greater acquisition values and the green 'X' represents the optimal point (0, 0).}
    \label{fig:visual}
\end{figure}

\subsubsection{Results and Analysis} \label{sec:db-res}
Figure \ref{fig:dmbs-res} (\textbf{Left}) summarizes the performance improvements of PostgreSQL on SysBench achieved by different algorithms. HiBO shows the highest throughput gains in greatest improvement rate, achieving about 28\% of performance improvements over the default performance (200 EPS) than LA-MCTS and TuRBO do in 50 iterations. Moreover, without additional pre-sampling, OtterTune$^*$ and LlamaTune also fall behind space-partitioning-based approaches. Such comparison highlights HiBO's effectiveness even in complex and noisy real-world scenarios.

Besides performance improvements, in real-world scenarios, tuning time and the rate of suggesting failure-causing configurations must be considered towards better practicality. A failure-causing configuration, defined as \textit{unsafe} \citep{zhang2022onlinetune}, can degrade the performance of critical DBMS-dependent processes in the system. Therefore, we introduce the \textbf{Safety-weighted Performance Improvement-Time Ratio} (\textbf{S-PITR}): $S\text{-}PITR = \frac{PI}{TT + NF \cdot PE}$,  where $PI$, $TT$, $NF$ and $PE$ refer to \textbf{p}erformance \textbf{i}mprovement, \textbf{t}uning \textbf{t}ime, \textbf{n}umber of \textbf{f}ailed (unsafe) configurations during the search and  the \textbf{pe}nalty value for failed configurations respectively. This metric quantitatively measures the practicality of tuning algorithms by considering both tuning time and safety.

Figure \ref{fig:dmbs-res} (\textbf{Right}) presents the S-PITR values for the selected algorithms while tuning PostgreSQL's configurations towards SysBench. Again, HiBO achieves the highest S-PITR value and outperforms LA-MCTS by over 20\%. Besides, HiBO presents $\sim$25\% greater S-PITR scores than OtterTune$^*$ and LlamaTune. These findings highlight HiBO's adequate ability to strike a balance between performance improvement, safety, and efficiency in complex real-world DBMS tuning problems.

\begin{figure}[t]
    \centering
    \includegraphics[scale=0.5]{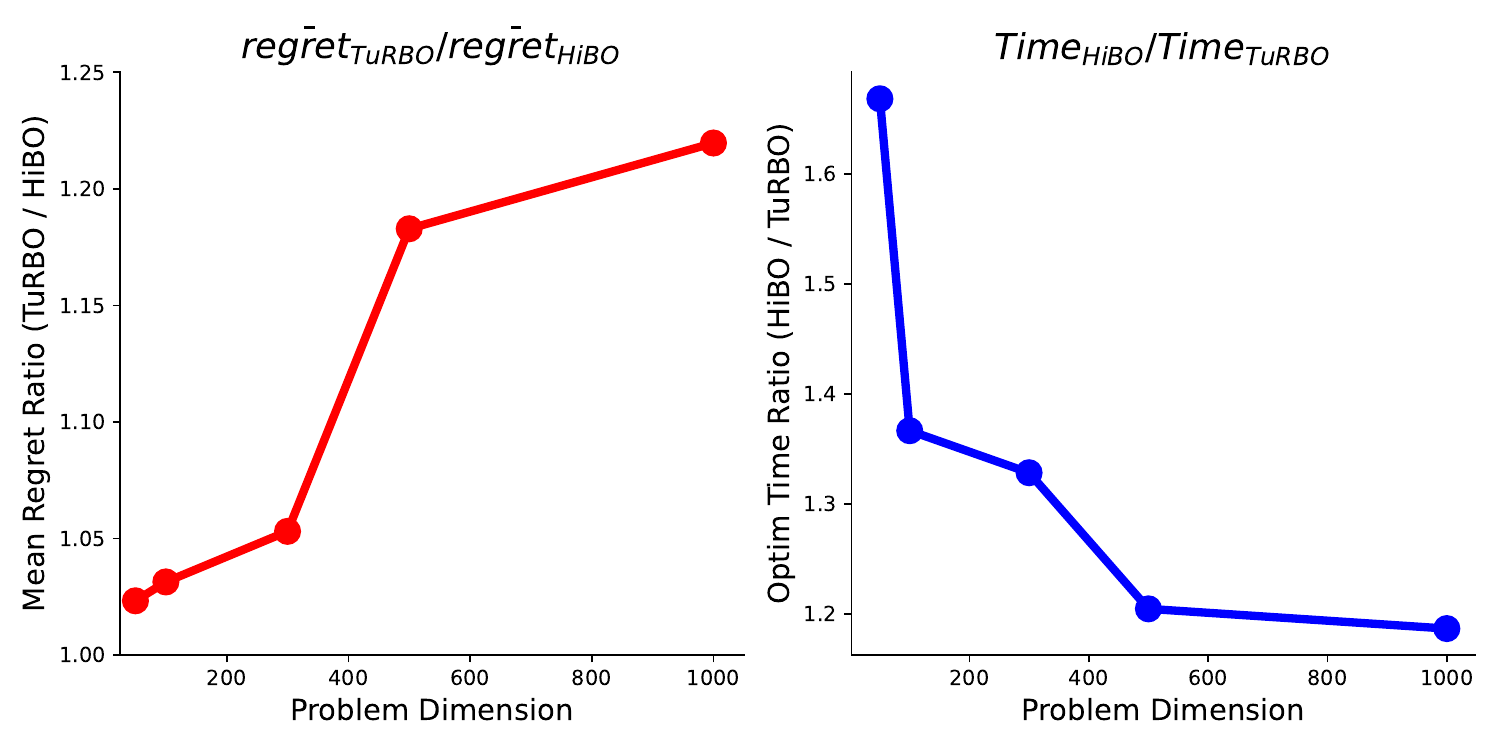}
    \caption{Comparison of HiBO and TuRBO across varying dimensionalities on the synthetic Levy benchmark. (\textbf{Left}) The ratio of mean regret (RMR) between TuRBO and HiBO ($\bar{regret}_{TuRBO} / \bar{regret}_{HiBO}$). (\textbf{Right}) The ratio of optimization time (ROT) for HiBO relative to TuRBO ($Time_{HiBO} / Time_{TuRBO}$).}
    \label{fig:levy_dim}
    \vskip -0.175in
\end{figure}

\subsection{Visual Analysis}
We visualize the acquisition function values generated by HiBO and TuRBO during the optimization of the Ackley2-200D synthetic function over 100 iterations. Starting with 20 initial samples, the distribution of acquisition values across the two effective dimensions as optimization progresses are plotted in Figure \ref{fig:visual}. For clarity, only the top 1000 acquisition values from 100x100 uniformly distributed samples are displayed at each iteration. The comparison shows that HiBO, leveraging space partitioning, converges its acquisition distribution more rapidly toward the optimal point, with the brightest areas near the optima. In contrast, TuRBO, while also moving toward the optimal region, converges more slowly and results in a less concentrated, noisier final distribution, with substantial acquisition values still scattered away from the optima. This analysis further validates the effectiveness of HiBO’s integration of space partitioning into the local optimizer’s acquisition strategy.

\subsection{Empirical Validation on Scalability}

To empirically evaluate HiBO's scalability with increasing dimensionality, we compare the performance and timing costs of HiBO against plain TuRBO on the synthetic Levy benchmark across varying dimensions, as shown in Figure \ref{fig:levy_dim}. Figure \ref{fig:levy_dim} (\textbf{Left}) depicts the ratio of mean regret (RMR) between TuRBO and HiBO ($\bar{regret}_{TuRBO} / \bar{regret}_{HiBO}$), which indicates HiBO’s performance advantage over TuRBO without the global-level navigator. It is observed that RMR increases steadily from 1.03 to 1.22 with dimensionality increasing from 50 to 1000, suggesting that TuRBO's mean regret on Levy grows more significantly than HiBO's with higher dimensionality, indicating that HiBO’s performance improvement scales effectively. 

Figure \ref{fig:levy_dim} (\textbf{Right}) illustrates the ratio of optimization time (ROT) for HiBO relative to TuRBO ($Time_{HiBO} / Time_{TuRBO}$), reflecting the efficiency overhead brought by HiBO’s global-level navigator. Although ROT remains above 1 due to the additional time cost brought by the global-level navigator of HiBO, ROT decreases from 1.66 to 1.18 as dimensionality increasing. This indicates that as dimensionality grows, the local optimizer's computational burden increases, while the global navigator's overhead becomes less significant. As a result, the global navigator’s relative overhead is less significant as the local optimizer becomes the dominant computational component at higher dimensions.

\section{Conclusion} In this paper, we introduce HiBO, a novel BO algorithm designed to efficiently handle high-dimensional search spaces via a hierarchical integration of global-level space partitioning and local modeling. Our comprehensive evaluation demonstrate its effectiveness and practical value on both synthetic benchmarks and real-world systems. Additionally, the visual analysis presents a straightforward validation of its effects in navigating on complex optimization landscapes. Its scalability w.r.t increased dimensionality is empirically validated on the synthetic benchmark. \textbf{More discussions on the scalability of HiBO can be found in Appendix \ref{eval:scalability}}.

\newpage
\bibliographystyle{unsrtnat}  
\bibliography{references}  

\begin{thebibliography}{54}
\providecommand{\natexlab}[1]{#1}
\providecommand{\url}[1]{\texttt{#1}}
\expandafter\ifx\csname urlstyle\endcsname\relax
  \providecommand{\doi}[1]{doi: #1}\else
  \providecommand{\doi}{doi: \begingroup \urlstyle{rm}\Url}\fi

\bibitem[Kandasamy et~al.(2018)Kandasamy, Neiswanger, Schneider, Poczos, and Xing]{kandasamy2018bo-nas}
Kirthevasan Kandasamy, Willie Neiswanger, Jeff Schneider, Barnabas Poczos, and Eric~P Xing.
\newblock Neural architecture search with bayesian optimisation and optimal transport.
\newblock \emph{Advances in neural information processing systems}, 31, 2018.

\bibitem[Ru et~al.(2020)Ru, Wan, Dong, and Osborne]{ru2020inter-bo}
Binxin Ru, Xingchen Wan, Xiaowen Dong, and Michael Osborne.
\newblock Interpretable neural architecture search via bayesian optimisation with weisfeiler-lehman kernels.
\newblock \emph{arXiv preprint arXiv:2006.07556}, 2020.

\bibitem[Hvarfner et~al.(2022)Hvarfner, Stoll, Souza, Lindauer, Hutter, and Nardi]{hvarfner2022pi-bo}
Carl Hvarfner, Danny Stoll, Artur Souza, Marius Lindauer, Frank Hutter, and Luigi Nardi.
\newblock $\pi${BO}: Augmenting acquisition functions with user beliefs for bayesian optimization.
\newblock \emph{arXiv preprint arXiv:2204.11051}, 2022.

\bibitem[Zimmer et~al.(2021)Zimmer, Lindauer, and Hutter]{zimmer2021auto-pytorch}
Lucas Zimmer, Marius Lindauer, and Frank Hutter.
\newblock Auto-pytorch: Multi-fidelity metalearning for efficient and robust autodl.
\newblock \emph{IEEE transactions on pattern analysis and machine intelligence}, 43\penalty0 (9):\penalty0 3079--3090, 2021.

\bibitem[Van~Aken et~al.(2017)Van~Aken, Pavlo, Gordon, and Zhang]{van2017otter}
Dana Van~Aken, Andrew Pavlo, Geoffrey~J Gordon, and Bohan Zhang.
\newblock Automatic database management system tuning through large-scale machine learning.
\newblock In \emph{Proceedings of the 2017 ACM international conference on management of data}, pages 1009--1024, 2017.

\bibitem[Zhang et~al.(2021)Zhang, Wu, Chang, Jin, Tan, Li, Zhang, and Cui]{zhang2021restune}
Xinyi Zhang, Hong Wu, Zhuo Chang, Shuowei Jin, Jian Tan, Feifei Li, Tieying Zhang, and Bin Cui.
\newblock Restune: Resource oriented tuning boosted by meta-learning for cloud databases.
\newblock In \emph{Proceedings of the 2021 international conference on management of data}, pages 2102--2114, 2021.

\bibitem[Zhang et~al.(2022)Zhang, Wu, Li, Tan, Li, and Cui]{zhang2022onlinetune}
Xinyi Zhang, Hong Wu, Yang Li, Jian Tan, Feifei Li, and Bin Cui.
\newblock Towards dynamic and safe configuration tuning for cloud databases.
\newblock In \emph{Proceedings of the 2022 International Conference on Management of Data}, pages 631--645, 2022.

\bibitem[Binois and Wycoff(2022)]{binois2022HDGPBO}
Mickael Binois and Nathan Wycoff.
\newblock A survey on high-dimensional gaussian process modeling with application to bayesian optimization.
\newblock \emph{ACM Transactions on Evolutionary Learning and Optimization}, 2\penalty0 (2):\penalty0 1--26, 2022.

\bibitem[Malu et~al.(2021)Malu, Dasarathy, and Spanias]{malu2021brief-HDBO}
Mohit Malu, Gautam Dasarathy, and Andreas Spanias.
\newblock Bayesian optimization in high-dimensional spaces: A brief survey.
\newblock In \emph{2021 12th International Conference on Information, Intelligence, Systems \& Applications (IISA)}, pages 1--8. IEEE, 2021.

\bibitem[Rashidi et~al.(2024)Rashidi, Johnstonbaugh, and Gao]{rashidi2024CTS}
Bahador Rashidi, Kerrick Johnstonbaugh, and Chao Gao.
\newblock Cylindrical thompson sampling for high-dimensional bayesian optimization.
\newblock In \emph{International Conference on Artificial Intelligence and Statistics}, pages 3502--3510. PMLR, 2024.

\bibitem[Zhao et~al.(2023)Zhao, Zhou, and Li]{zhao2023dbms-survey}
Xinyang Zhao, Xuanhe Zhou, and Guoliang Li.
\newblock Automatic database knob tuning: a survey.
\newblock \emph{IEEE Transactions on Knowledge and Data Engineering}, 2023.

\bibitem[Cereda et~al.(2021)Cereda, Valladares, Cremonesi, and Doni]{cereda2021cgptuner}
Stefano Cereda, Stefano Valladares, Paolo Cremonesi, and Stefano Doni.
\newblock Cgptuner: a contextual gaussian process bandit approach for the automatic tuning of it configurations under varying workload conditions.
\newblock \emph{Proceedings of the VLDB Endowment}, 14\penalty0 (8):\penalty0 1401--1413, 2021.

\bibitem[Eriksson et~al.(2019)Eriksson, Pearce, Gardner, Turner, and Poloczek]{eriksson2019turbo}
David Eriksson, Michael Pearce, Jacob Gardner, Ryan~D Turner, and Matthias Poloczek.
\newblock Scalable global optimization via local bayesian optimization.
\newblock \emph{Advances in neural information processing systems}, 32, 2019.

\bibitem[Daulton et~al.(2022)Daulton, Eriksson, Balandat, and Bakshy]{daulton2022MORBO}
Samuel Daulton, David Eriksson, Maximilian Balandat, and Eytan Bakshy.
\newblock Multi-objective bayesian optimization over high-dimensional search spaces.
\newblock In \emph{Uncertainty in Artificial Intelligence}, pages 507--517. PMLR, 2022.

\bibitem[Wang et~al.(2020)Wang, Fonseca, and Tian]{wang2020lamcts}
Linnan Wang, Rodrigo Fonseca, and Yuandong Tian.
\newblock Learning search space partition for black-box optimization using monte carlo tree search.
\newblock \emph{Advances in Neural Information Processing Systems}, 33:\penalty0 19511--19522, 2020.

\bibitem[Munos(2011)]{munos2011doo}
R{\'e}mi Munos.
\newblock Optimistic optimization of a deterministic function without the knowledge of its smoothness.
\newblock \emph{Advances in neural information processing systems}, 24, 2011.

\bibitem[Bubeck et~al.(2011)Bubeck, Munos, Stoltz, and Szepesv{\'a}ri]{bubeck2011hoo}
S{\'e}bastien Bubeck, R{\'e}mi Munos, Gilles Stoltz, and Csaba Szepesv{\'a}ri.
\newblock X-armed bandits.
\newblock \emph{Journal of Machine Learning Research}, 12\penalty0 (5), 2011.

\bibitem[Kim et~al.(2020)Kim, Lee, Lim, Kaelbling, and Lozano-P{\'e}rez]{kim2020voo}
Beomjoon Kim, Kyungjae Lee, Sungbin Lim, Leslie Kaelbling, and Tom{\'a}s Lozano-P{\'e}rez.
\newblock Monte carlo tree search in continuous spaces using voronoi optimistic optimization with regret bounds.
\newblock In \emph{Proceedings of the AAAI Conference on Artificial Intelligence}, volume~34, pages 9916--9924, 2020.

\bibitem[Jones(2008)]{jones2008mopta}
Donald~R Jones.
\newblock Large-scale multi-disciplinary mass optimization in the auto industry.
\newblock In \emph{MOPTA 2008 Conference (20 August 2008)}, volume~64, 2008.

\bibitem[Wang et~al.(2017)Wang, Li, Jegelka, and Kohli]{wang2017rover}
Zi~Wang, Chengtao Li, Stefanie Jegelka, and Pushmeet Kohli.
\newblock Batched high-dimensional bayesian optimization via structural kernel learning.
\newblock In \emph{International conference on machine learning}, pages 3656--3664. PMLR, 2017.

\bibitem[Candelieri(2021)]{candelieri2021gentle-bo}
Antonio Candelieri.
\newblock A gentle introduction to bayesian optimization.
\newblock In \emph{2021 Winter Simulation Conference (WSC)}, pages 1--16. IEEE, 2021.

\bibitem[Siivola et~al.(2018)Siivola, Vehtari, Vanhatalo, Gonz{\'a}lez, and Andersen]{siivola2018correcting}
Eero Siivola, Aki Vehtari, Jarno Vanhatalo, Javier Gonz{\'a}lez, and Michael~Riis Andersen.
\newblock Correcting boundary over-exploration deficiencies in bayesian optimization with virtual derivative sign observations.
\newblock In \emph{2018 IEEE 28th International Workshop on Machine Learning for Signal Processing (MLSP)}, pages 1--6. IEEE, 2018.

\bibitem[Duvenaud et~al.(2011)Duvenaud, Nickisch, and Rasmussen]{duvenaud2011additiveGP}
David~K Duvenaud, Hannes Nickisch, and Carl Rasmussen.
\newblock Additive gaussian processes.
\newblock \emph{Advances in neural information processing systems}, 24, 2011.

\bibitem[Kandasamy et~al.(2015)Kandasamy, Schneider, and P{\'o}czos]{kandasamy2015high-add}
Kirthevasan Kandasamy, Jeff Schneider, and Barnab{\'a}s P{\'o}czos.
\newblock High dimensional bayesian optimisation and bandits via additive models.
\newblock In \emph{International conference on machine learning}, pages 295--304. PMLR, 2015.

\bibitem[Wang et~al.(2018)Wang, Gehring, Kohli, and Jegelka]{wang2018ebo}
Zi~Wang, Clement Gehring, Pushmeet Kohli, and Stefanie Jegelka.
\newblock Batched large-scale bayesian optimization in high-dimensional spaces.
\newblock In \emph{International Conference on Artificial Intelligence and Statistics}, pages 745--754. PMLR, 2018.

\bibitem[Wang et~al.(2016)Wang, Hutter, Zoghi, Matheson, and De~Feitas]{wang2016rembo}
Ziyu Wang, Frank Hutter, Masrour Zoghi, David Matheson, and Nando De~Feitas.
\newblock Bayesian optimization in a billion dimensions via random embeddings.
\newblock \emph{Journal of Artificial Intelligence Research}, 55:\penalty0 361--387, 2016.

\bibitem[Letham et~al.(2020)Letham, Calandra, Rai, and Bakshy]{letham2020re}
Ben Letham, Roberto Calandra, Akshara Rai, and Eytan Bakshy.
\newblock Re-examining linear embeddings for high-dimensional bayesian optimization.
\newblock \emph{Advances in neural information processing systems}, 33:\penalty0 1546--1558, 2020.

\bibitem[Nayebi et~al.(2019)Nayebi, Munteanu, and Poloczek]{nayebi2019hesbo}
Amin Nayebi, Alexander Munteanu, and Matthias Poloczek.
\newblock A framework for bayesian optimization in embedded subspaces.
\newblock In \emph{International Conference on Machine Learning}, pages 4752--4761. PMLR, 2019.

\bibitem[Papenmeier et~al.(2022)Papenmeier, Nardi, and Poloczek]{papenmeier2022baxus}
Leonard Papenmeier, Luigi Nardi, and Matthias Poloczek.
\newblock Increasing the scope as you learn: Adaptive bayesian optimization in nested subspaces.
\newblock \emph{Advances in Neural Information Processing Systems}, 35:\penalty0 11586--11601, 2022.

\bibitem[Li et~al.(2017)Li, Gupta, Rana, Nguyen, Venkatesh, and Shilton]{li2017dropoutBO}
Cheng Li, Sunil Gupta, Santu Rana, Vu~Nguyen, Svetha Venkatesh, and Alistair Shilton.
\newblock High dimensional bayesian optimization using dropout.
\newblock In \emph{Proceedings of the 26th International Joint Conference on Artificial Intelligence}, IJCAI'17, page 2096–2102. AAAI Press, 2017.
\newblock ISBN 9780999241103.

\bibitem[Spagnol et~al.(2019)Spagnol, Le~Riche, and da~Veiga]{spagnol19hsic}
Adrien Spagnol, Rodolphe Le~Riche, and S{\'e}bastien da~Veiga.
\newblock {Bayesian optimization in effective dimensions via kernel-based sensitivity indices}.
\newblock In \emph{{13th International Conference on Applications of Statistics and Probability in Civil Engineering(ICASP13)}}, S{\'e}oul, South Korea, May 2019.
\newblock \doi{10.22725/ICASP13.093}.
\newblock URL \url{https://hal-emse.ccsd.cnrs.fr/emse-02133923}.

\bibitem[Shen and Kingsford(2023)]{shen2023vsbo}
Yihang Shen and Carl Kingsford.
\newblock Computationally efficient high-dimensional bayesian optimization via variable selection.
\newblock In \emph{AutoML Conference 2023}, 2023.

\bibitem[Song et~al.(2022)Song, Xue, Huang, and Qian]{song2022mctsvs}
Lei Song, Ke~Xue, Xiaobin Huang, and Chao Qian.
\newblock Monte carlo tree search based variable selection for high dimensional bayesian optimization.
\newblock \emph{Advances in Neural Information Processing Systems}, 35:\penalty0 28488--28501, 2022.

\bibitem[Wang et~al.(2021)Wang, Xie, Li, Fonseca, and Tian]{wang2021la-nas}
Linnan Wang, Saining Xie, Teng Li, Rodrigo Fonseca, and Yuandong Tian.
\newblock Sample-efficient neural architecture search by learning actions for monte carlo tree search.
\newblock \emph{IEEE Transactions on Pattern Analysis and Machine Intelligence}, 44\penalty0 (9):\penalty0 5503--5515, 2021.

\bibitem[MacQueen et~al.(1967)]{macqueen1967kmeans}
James MacQueen et~al.
\newblock Some methods for classification and analysis of multivariate observations.
\newblock In \emph{Proceedings of the fifth Berkeley symposium on mathematical statistics and probability}, volume~1, pages 281--297. Oakland, CA, USA, 1967.

\bibitem[Kocsis and Szepesv{\'a}ri(2006)]{kocsis2006UCT}
Levente Kocsis and Csaba Szepesv{\'a}ri.
\newblock Bandit based monte-carlo planning.
\newblock In \emph{European conference on machine learning}, pages 282--293. Springer, 2006.

\bibitem[Boser et~al.(1992)Boser, Guyon, and Vapnik]{boser1992svm}
Bernhard~E Boser, Isabelle~M Guyon, and Vladimir~N Vapnik.
\newblock A training algorithm for optimal margin classifiers.
\newblock In \emph{Proceedings of the fifth annual workshop on Computational learning theory}, pages 144--152, 1992.

\bibitem[Hinton et~al.(2015)Hinton, Vinyals, and Dean]{hinton2015distilling}
Geoffrey Hinton, Oriol Vinyals, and Jeff Dean.
\newblock Distilling the knowledge in a neural network.
\newblock \emph{arXiv preprint arXiv:1503.02531}, 2015.

\bibitem[Lozano et~al.(2006)Lozano, Larra{\~n}aga, Inza, and Bengoetxea]{lozano2006cma-es}
Jose~A Lozano, Pedro Larra{\~n}aga, I{\~n}aki Inza, and Endika Bengoetxea.
\newblock \emph{Towards a new evolutionary computation: advances on estimation of distribution algorithms}, volume 192.
\newblock Springer, 2006.

\bibitem[Eriksson and Jankowiak(2021)]{eriksson2021saasbo}
David Eriksson and Martin Jankowiak.
\newblock High-dimensional bayesian optimization with sparse axis-aligned subspaces.
\newblock In \emph{Uncertainty in Artificial Intelligence}, pages 493--503. PMLR, 2021.

\bibitem[{OnGres Inc.}(2023)]{PostgreSQLCO.NF}
{OnGres Inc.}
\newblock Postgresqlco.nf.
\newblock \url{https://postgresqlco.nf/doc/en/param/}, 2023.

\bibitem[Kopytov(2004)]{kopytov2004sysbench}
Alexey Kopytov.
\newblock Sysbench: a system performance benchmark.
\newblock \emph{http://sysbench. sourceforge. net/}, 2004.

\bibitem[Kanellis et~al.(2022)Kanellis, Ding, Kroth, M{\"u}ller, Curino, and Venkataraman]{kanellis2022llamatune}
Konstantinos Kanellis, Cong Ding, Brian Kroth, Andreas M{\"u}ller, Carlo Curino, and Shivaram Venkataraman.
\newblock Llamatune: Sample-efficient dbms configuration tuning.
\newblock \emph{arXiv preprint arXiv:2203.05128}, 2022.

\bibitem[Pedregosa et~al.(2011)Pedregosa, Varoquaux, Gramfort, Michel, Thirion, Grisel, Blondel, Prettenhofer, Weiss, Dubourg, et~al.]{pedregosa2011scikit}
Fabian Pedregosa, Ga{\"e}l Varoquaux, Alexandre Gramfort, Vincent Michel, Bertrand Thirion, Olivier Grisel, Mathieu Blondel, Peter Prettenhofer, Ron Weiss, Vincent Dubourg, et~al.
\newblock Scikit-learn: Machine learning in python.
\newblock \emph{the Journal of machine Learning research}, 12:\penalty0 2825--2830, 2011.

\bibitem[Mat{\'e}rn(2013)]{matern2013matern}
Bertil Mat{\'e}rn.
\newblock \emph{Spatial variation}, volume~36.
\newblock Springer Science \& Business Media, 2013.

\bibitem[Jones et~al.(1998)Jones, Schonlau, and Welch]{jones1998EI}
Donald~R Jones, Matthias Schonlau, and William~J Welch.
\newblock Efficient global optimization of expensive black-box functions.
\newblock \emph{Journal of Global optimization}, 13:\penalty0 455--492, 1998.

\bibitem[Hansen et~al.(2019)Hansen, Akimoto, and Baudis]{hansen2019pycma}
Nikolaus Hansen, Youhei Akimoto, and Petr Baudis.
\newblock {CMA-ES/pycma} on {G}ithub.
\newblock Zenodo, DOI:10.5281/zenodo.2559634, February 2019.
\newblock URL \url{https://doi.org/10.5281/zenodo.2559634}.

\bibitem[Balandat et~al.(2020)Balandat, Karrer, Jiang, Daulton, Letham, Wilson, and Bakshy]{balandat2020botorch}
Maximilian Balandat, Brian Karrer, Daniel Jiang, Samuel Daulton, Ben Letham, Andrew~G Wilson, and Eytan Bakshy.
\newblock Botorch: A framework for efficient monte-carlo bayesian optimization.
\newblock \emph{Advances in neural information processing systems}, 33:\penalty0 21524--21538, 2020.

\bibitem[Neal(2012)]{neal2012ARD}
Radford~M Neal.
\newblock \emph{Bayesian learning for neural networks}, volume 118.
\newblock Springer Science \& Business Media, 2012.

\bibitem[Thompson(1933)]{thompson1933TS}
William~R Thompson.
\newblock On the likelihood that one unknown probability exceeds another in view of the evidence of two samples.
\newblock \emph{Biometrika}, 25\penalty0 (3-4):\penalty0 285--294, 1933.

\bibitem[Vert et~al.(2004)Vert, Tsuda, and Sch{\"o}lkopf]{vert2004kernel}
Jean-Philippe Vert, Koji Tsuda, and Bernhard Sch{\"o}lkopf.
\newblock A primer on kernel methods.
\newblock 2004.

\bibitem[Lindauer et~al.(2022)Lindauer, Eggensperger, Feurer, Biedenkapp, Deng, Benjamins, Ruhkopf, Sass, and Hutter]{lindauer2022smac3}
Marius Lindauer, Katharina Eggensperger, Matthias Feurer, Andr{\'e} Biedenkapp, Difan Deng, Carolin Benjamins, Tim Ruhkopf, Ren{\'e} Sass, and Frank Hutter.
\newblock Smac3: A versatile bayesian optimization package for hyperparameter optimization.
\newblock \emph{Journal of Machine Learning Research}, 23\penalty0 (54):\penalty0 1--9, 2022.

\bibitem[{PostgreSQL Global Development Group}(2023)]{PostgreSQL}
{PostgreSQL Global Development Group}.
\newblock Postgresql.
\newblock \url{https://www.postgresql.org/docs/16/index.html}, 2023.

\bibitem[{Merit Data Tech}(2023)]{whyPGSQL}
{Merit Data Tech}.
\newblock Why data engineers use postgresql for large, data-driven applications.
\newblock \url{https://www.meritdata-tech.com/resources/blog/digital-engineering-solutions/postgresql-large-data-applications/}, 2023.

\end{thebibliography}
\newpage
\appendix
\onecolumn
\section{Experiment Settings}\label{app: expr_set}
\subsection{Algorithm Configurations for Synthetic Benchmarks}\label{app:syn-config}
\begin{itemize}
    \item \textbf{Random Search}: Uniform sampling using function `np.random.uniform`\footnote{https://numpy.org/doc/stable/reference/random/generated/numpy.random.uniform.html}
    \item \textbf{GP-BO}: Implemented using `GaussianProcessRegressor` from scikit-learn library \citep{pedregosa2011scikit},  configured with a Matérn kernel ($\nu=2.5$) \citep{matern2013matern} scaled by a constant factor of 1.0, and the noise level is set to 0.1, which translates to an alpha value of 0.01. The acquisition function is set to Expected Improvement \citep{jones1998EI};
    \item \textbf{CMA-ES}: We used the out-of-box PyCMA \citep{hansen2019pycma} implementation of CMA-ES, where initial standard deviation is set to 0.5;
    \item \textbf{TuRBO}: hyperparameter settings are divided into two parts: 1) about the GP model: implemented using BoTorch \citep{balandat2020botorch} with noise constraint interval of $[1 \times 10^{-8}, 1 \times 10^{-3}]$, a scaled Matérn kernel ($\nu = 2.5$) and Automatic Relevance Determination (ARD; \citet{neal2012ARD}) lengthscales for each input dimension constrained within $[0.005, 4.0]$; 2) TuRBO-related: thresholds of consecutive successes and failures to trigger trust region size changing are set to 3 and 5 respectively, with the minimum trust region length being 0.03125 within normalized range. The acquisition function is set to Thompson Sampling (TS; \citet{thompson1933TS}.
    \item \textbf{LA-MCTS}: Use TuRBO as the local model with the same configuration mentioned above for a separate TuRBO. The leaf size (split threshold) is set to 10 samples. The kernel type for SVM used in its MCTS is radial basis function (RBF; \citet{vert2004kernel})
    \item \textbf{HiBO}: Temperature for Softmax is set to 0.1. The limit of maximum tree depth is 5. $C_p$ is set to 0.5.  Thresholds of consecutive successes and failures to trigger trust region size changing are set to 5 and 3 respectively. Use TuRBO as the local model with the same configuration mentioned above for a separate TuRBO. Use the same SVM classifier for search-tree-based space partitioning in the global-level navigator.
\end{itemize}

\subsection{Experiment Setup for DBMS Configuration Tuning}\label{app:dbms-config}
The algorithms used in both synthetic and DBMS configuration tuning have the same configurations as they have in synthetic benchmark evaluation. The following are configurations for the two algorithms specifically designed for DBMS configuration tuning.

\begin{itemize}
\setlength\itemsep{-0.2em}
    \item \textbf{OtterTune*}: As mentioned in Section \ref{sec:dbms-setup}, this is a simplified version of OtterTune \citep{van2017otter} without workload mapping and statistical knob selection modules. It applies the vanilla GP-based BO for tuning in our setting to fit the scenario with very limited sample budget and high dimensionalities. The configurations for GP follow those used in synthetic benchmarks.
    \item \textbf{LlamaTune}: LlamaTune \citep{kanellis2022llamatune} is the most relevant work to ours in the context of DBMS configuration tuning. It directly tunes the DBMS over the high-dimensional configuration space with limited prior knowledge, employing subspace-embedding methods \citep{nayebi2019hesbo, wang2016rembo} to scale vanilla BO to high-dimensional tasks. We follow the experimental setup specified in Section 6.1 of their paper, including the use of HeSBO \citep{nayebi2019hesbo} random projections with 16 dimensions, a 20\% biasing towards special values, and bucketizing the search space by limiting each dimension to 10,000 unique values. The base surrogate model is SMAC, implemented using the SMAC3 library \citep{lindauer2022smac3}.
\end{itemize}

The experiments for DBMS configuration tuning were conducted on two VM instances with identical configurations. Each of them is equipped with 8 single-core Intel(R) Xeon(R) Gold 6142 CPU @ 2.60GHz, 64GB RAM and 256GB SSD. One instance is used for executing tuning management and optimization algorithm logic. The DBMS and stored databases are stored in the other instance. One instance leverages SSH to control the other for executing DBMS workloads. 

We select PostgreSQL v16.3 \citep{PostgreSQL} considering its accessibility and generality. PostgreSQL is a powerful, open-source Relational DBMS that has become one of the most popular choices for managing enterprise-level databases \citep{whyPGSQL} given its reliability and extensibility. Importantly, PostgreSQL exposes up to more than 200 configuration knobs as shown in PGCONF \citep{PostgreSQLCO.NF} which makes it an ideal instance for our DBMS configuration tuning experiments.  We collected information of all configuration knobs of PostgreSQL in PostgreSQL Co.NF \citep{PostgreSQLCO.NF}. A total of \textbf{110} knobs were manually selected, excluding those related to debugging, network connections, and authorization, as they could interfere with the interaction between the DBMS controller module and the DBMS instance and have negligible impact on DBMS performance.

To simulate the limited sample budget typical in production environments, we use only 50 iterations for tuning, with the initial 10 iterations reserved for random sampling. This is also based on the observation that tuning for more iterations brings no significant improvement while resulting in extra tuning time overhead. Each iteration involves a 100-second workload execution. Configurations that fail to launch PostgreSQL or cause execution errors during execution are penalized with a throughput value of zero.

\subsection{Reasoning: Algorithm Selection in DBMS Configuration Tuning Experiments}

This section provides the clarification of our selection on BO-based approaches specifically for DMBS configuration tuning. Many BO-based methods in this field \citep{zhang2021restune, zhang2022onlinetune, cereda2021cgptuner} and the full-version OtterTune were not included in our evaluation due to the inapplicability of their special data and different focus:
\begin{itemize}
\setlength\itemsep{-0.16em}
    \item Some methods rely on pre-collected data or additional knowledge before actual tuning as mentioned in Section \ref{sec:intro}. For example, OtterTune requires a large number of random samples to be collected, while ResTune \citep{zhang2021restune} applies workload features and observations from 34 prior tuning tasks. These requirements add substantial overhead and are impractical for direct comparison, and reproducing their setups is infeasible without the required data. Our experiments show that HiBO achieves significant throughput improvements on PostgreSQL without pre-collected data and within a limited sample budget, providing an implicit comparison of its effectiveness even without directly implementing these methods.
    \item Moreover, these works focus on low-dimensional tasks rather than directly optimizing high-dimensional configuration spaces. OtterTune selects important knobs for dimensionality reduction through pre-tuning sampling, while  ResTune, OnlineTune \citep{zhang2022onlinetune} and CGPTuner \citep{cereda2021cgptuner} limit their evaluations to using fewer than 50 knobs through manual selection. In contrast, HiBO directly tunes the high-dimensional configuration space without requiring dimensionality reduction, prior knowledge, or complex models, while maintaining high sample efficiency. 
\end{itemize}
Therefore, we only introduced the partial version of OtterTune as a baseline (OtterTune$^*$) and LlamaTune, which also applies high-dimensional BO methods.

\subsection{SysBench Statement Types}\label{app: sys-code}
We used the script `oltp\_read\_write.lua` released by the authors of SysBench \citep{kopytov2004sysbench} for benchmarking, which contains the following types of queries:
\begin{itemize}
\setlength
  \item Point selection: \texttt{"SELECT c FROM sbtest\%u WHERE id=?"}
  \item Range selection: \texttt{"SELECT c FROM sbtest\%u WHERE id BETWEEN ? AND ?"}
  \item Range selection with sum: \texttt{"SELECT SUM(k) FROM sbtest\%u WHERE id BETWEEN ? AND ?"}
  \item Range selection with ordering: \texttt{"SELECT c FROM sbtest\%u WHERE id BETWEEN ? AND ? ORDER BY c"}
  \item Range selection with distinction: \texttt{"SELECT DISTINCT c FROM sbtest\%u WHERE id BETWEEN ? AND ? ORDER BY c"}
  \item Index update: \texttt{"UPDATE sbtest\%u SET k=k+1 WHERE id=?"}
  \item Non-index update: \texttt{"UPDATE sbtest\%u SET c=? WHERE id=?"}
  \item Deletion: \texttt{"DELETE FROM sbtest\%u WHERE id=?"}
  \item Insertion: \texttt{"INSERT INTO sbtest\%u (id, k, c, pad) VALUES (?, ?, ?, ?)"}
\end{itemize}
where "?"  here stands for parameters to be automatically specified by SysBench when these queries are executed.

\subsection{Hyperparameters} \label{app: hyper}


\begin{table}[t]
\vskip 0.15in
\begin{center}
\begin{small}

\bgroup
\def\arraystretch{1.3}%
\begin{tabular}{p{3.8cm}p{7.7cm}p{1.4cm}}
\toprule
\textbf{Hyperparameter} & \textbf{Description} & \textbf{Range} \\ 
\midrule
$C_p$ (UCT Constant) & Balances exploration and exploitation in the UCT formula. & [0.5, 5.0] \\

$\tau$ (Softmax Temperature) & Controls the impact of global-level navigator's guidance by adjusting the smoothness of sampling potential distribution. & [0.01, 2.0]  \\ 
Maximum Tree Depth Limit & Threshold of maximum tree depth, beyond which a restart should happen after consecutive failures. & [3, 5] \\ 
\bottomrule
\end{tabular}
\egroup

\end{small}
\end{center}
\caption{Hyperparameters in HiBO.}
\label{tab:hibo_hyperparameters}
\vspace{-10pt}
\end{table}

This section summarizes the hyperparameters used in our work, including their purposes and the ranges employed during grid search to determine the optimal combination. The specific setting of algorithms used in experiments can be found in Appendix \ref{app:syn-config} and Appendix \ref{app:dbms-config}, while the ranges presented here are those we used for grid search to find the best parameter value combination:
\begin{itemize}
\setlength\itemsep{-0.16em}
    \item $C_p$: Controls the balance between exploration and exploitation in the UCT calculation (Equation \ref{eq:UCT}). Larger values increase the weight of exploration. Range: [0.5, 5.0].
    \item $\tau$: Softmax temperature for normalizing partition scores into a positive and normalized range. Detailed interpretation and ablation studies about this hyperparameter are available in Appendix \ref{app: temp}. Range: [0.01, 2].
    \item \textbf{Maximum Tree Depth Limit}: Sets the limit of maximum allowable tree depth before the search process restarts after multiple times of repeated failures. It controls the tolerance of consecutive failures in one search. Range: [3, 5].
\end{itemize}

A naive grid search was used to identify the optimal combination of these hyperparameters within the specified ranges. Table \ref{tab:hibo_hyperparameters} provides a concise summary for reference.

\section{Theoretical Analysis}\label{app:add_theory}

\subsection{Preliminaries and Assumption}

\begin{assumption}
\label{assump:density}
\textbf{(Finite Sample Domain \& Performance Distribution).}
Let $\Omega$ be a finite search domain of size $N$, e.g., $N$ candidate solutions (or architectures).  Suppose there exists a probability density function $f(v)$ over the (unknown) performance $v \in [0,1]$.  Then for any $0 \le a < b \le 1$, we have:
\(
P(a < v < b) \;=\; \int_{a}^{b} f(v)\, dv.
\)
Additionally, assume the standard deviation $\sigma_v$ of $v$ is finite, and let $\bar{v}$ and $M_v$ denote the mean and median of $v$ (w.r.t.\ $f$), respectively.  Then, we have
\begin{equation}
\label{eq:mean-med-bound}
\bigl|\mathbb{E}(\bar{v} - M_v)\bigr| \;<\; \sigma_v.
\end{equation}
\end{assumption}
This assumption allows us to bound the fraction of solutions whose performance lies near the median or near the mean.  It also induces an upper bound on how unevenly $\Omega$ can be split according to performance metrics.

\subsection{Proof of Theorem~\ref{thm:leaf-bound}} \label{app: proof_them}

\begin{proof}
Let $\delta_1, \delta_2, \ldots, \delta_h$ be the partition errors at successive tree levels along the path to the leftmost leaf.  At the root, we split $\Omega$ into two sets of size at most $N/2 + \delta_1$ and $N/2 - \delta_1$, respectively.  By assumption, the \emph{left child} takes the larger set, hence it has size 
\[
N/2 \;+\; \delta_1.
\]
At the next level, we apply the same argument to that child’s sub-domain of size $N/2 + \delta_1$, resulting in two further splits.  In the worst case, the sub-domain with the highest partition score again takes the larger portion: 
\[
\frac{N}{2} \;+\; \delta_1 \;\;\longrightarrow\;\; \frac{N}{2^2} \;+\; \frac{\delta_1}{2} \;+\; \delta_2 
\quad\text{(at level 2)},
\]
and so on, possibly incurring a partition error $\delta_j$ at each stage.  Recursively unrolling this for $h$ splits, we get
\[
\underbrace{\delta_h \;+\; \frac{\delta_{h-1}}{2} \;+\; \frac{\delta_{h-2}}{2^2} \;+\;\dots \;+\; \frac{\delta_1}{2^{h-1}}}_{\text{accumulated partition error}}
\;+\;\frac{N}{2^h}.
\]
Defining $\delta_{\max} = \max\{\delta_1,\delta_2,\dots,\delta_h\}$ and bounding each $\delta_j$ by $\delta_{\max}$ yields:
\begin{align*}
\delta_h + \frac{\delta_{h-1}}{2} + \dots + \frac{\delta_1}{2^{h-1}}
&\;\;\le\;\;
\delta_{\max}\,\Bigl(1 \;\;+\dots\;\frac{1}{2^{\,h-1}}\Bigr) \\
&\;=\;
2\,\delta_{\max}\,\Bigl(1 - \frac{1}{2^h}\Bigr).
\end{align*}
Combining this with $\tfrac{N}{2^h}$ completes the proof,
\[
\underbrace{\delta_h + \dots + \frac{\delta_1}{2^{h-1}}}_{\le\,2\,\delta_{\max}\left(1 - \tfrac{1}{2^h}\right)}
\;+\;\frac{N}{2^h}
\;\;\le\;\;
2\,\delta_{\max}\,\Bigl(1 - \frac{1}{2^h}\Bigr)\;+\;\frac{N}{2^h}.
\]
\end{proof}

\subsection{Comparison to LA-NAS and LA-MCTS}
\label{sec:compare-lanas-lamcts}

\paragraph{Single-Path vs.\ Weighted Sampling.}
Methods like \textbf{LA-NAS}~\citep{wang2021la-nas} and \textbf{LA-MCTS}~\citep{wang2020lamcts} also partition the search space but only sample from the estimated most promising partition. For example, in LA-NAS’s theoretical analysis, all subsequent splits are localized to that single path, so that the domain of interest gets halved (plus partition error) at each level. This provides an exponential shrinkage bound (similar to Theorem~\ref{thm:leaf-bound}) for that single partition.

\paragraph{HiBO’s Approach: Partition-Score Weighting.}
By contrast, \textbf{HiBO} leverages the partition scores of all resulting leaf nodes for weighing the acquisition function values, instead of discarding sampling potential information from other partitions. This addresses a key limitation of single-path methods: \emph{if early splits are misleading}, LA-NAS (or LA-MCTS) can end up discarding regions that later prove to contain globally better optima.  In HiBO:
\begin{itemize}\setlength\itemsep{-0.16em}
    \item Leaves in which the local surrogate (and UCT) reveal good performance retain a high value of acquisition function.
    \item Leaves with poor performance or high uncertainty gradually lose weight but are still revisited occasionally, mitigating the risk of “missing” hidden optima.
\end{itemize}


\subsection{Scalability Analysis}\label{eval:scalability}
An analysis about how the search-tree-based space partitioning scales with the dimensionality $d$ of the search space and the total number of samples $T$ is given below.


\textbf{Dimension vs.\ Samples.} For continuous or high-dimensional discrete $\Omega$, we typically let $N$ grow exponentially in $d$ (e.g., a grid with spacing $\Delta$ leads to $\text{grid size} \approx (1/\Delta)^d$).  
In such scenarios, \textbf{any} method that enumerates or partitions $\Omega$ \emph{naively} could face exponential blow-up in $d$.  
HiBO mitigates this by:
\begin{enumerate}
    \item \textbf{Adaptive Expansion:} Partitions are grown incrementally; subspaces with low potential (low UCT scores) are not expanded repeatedly.
    \item \textbf{Weighted Sampling:} Unlike methods that only focus on a single path, HiBO keeps a global perspective to hedge against misleading early splits; yet high-performing partitions do receive exponentially more attention, so the search zooms in on promising regions quickly.
\end{enumerate}
Consequently, while worst-case complexity may still be large for extremely high $d$, in practice HiBO often scales better than vanilla BO with full GP modeling, which suffers from $O(T^3)$ kernel-inversion costs or from poor kernel fit in high dimensions.

\section{Ablation Study: Adaptive Maximum Tree Depth} \label{app: ada-depth}
\begin{figure}[h]
    \centering
    \includegraphics[width=\linewidth]{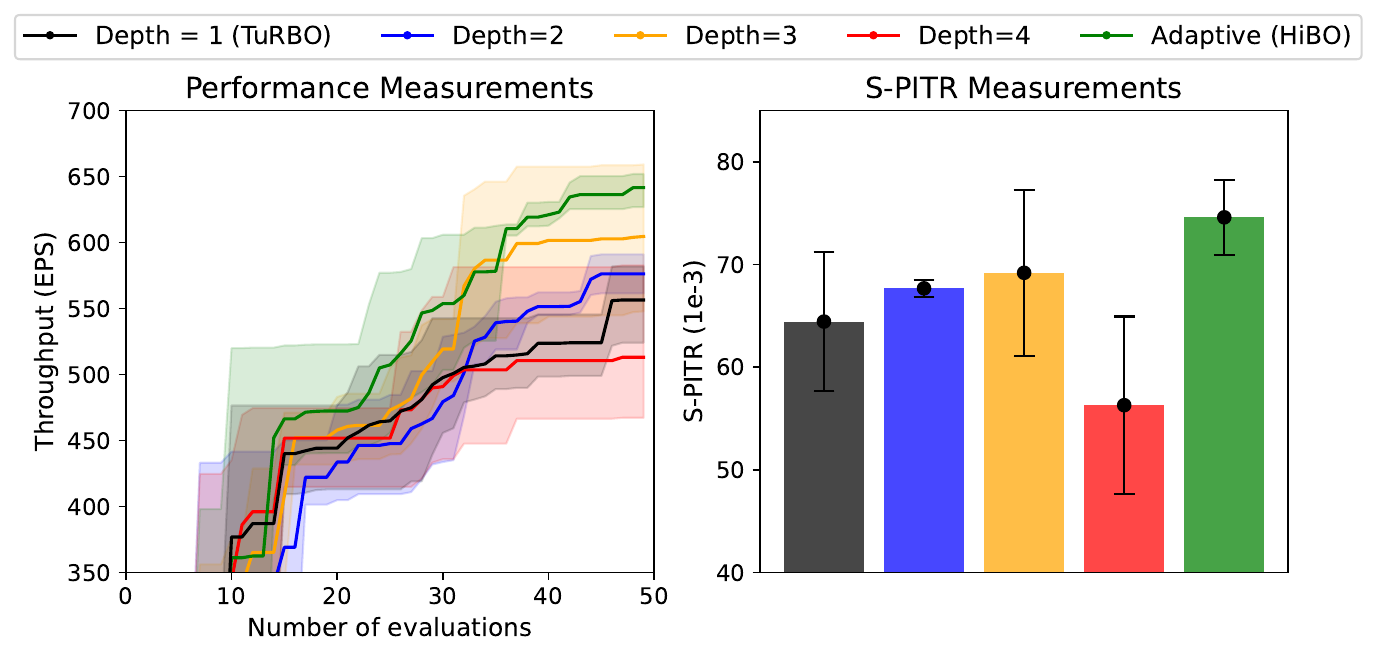}
    \caption{Performance improvement evaluation and measurements of S-PITR for HiBO with different maximum tree depth settings on SysBench.}
    \label{fig:abs_depth}
\end{figure}

In this section, we conduct an ablation study on the adaptively adjusted maximum depth of the search tree to validate its effects. As discussed in Section \ref{sec:ssp-rules}, we designed extra rules to dynamically adjust the maximum tree depth for the search tree depending on consecutive successes and failures, aiming to 1) reduce unnecessary computational cost and 2) balance exploitation and exploration by adjusting the range covered by resulting search space partitions. To assess the effects of these designs, we evaluate HiBO under different tree depth settings, considering both efficiency and computational cost.

For this experiment, we use DBMS configuration tuning with workload being SysBench as the benchmark, following the same configuration setup described in Section \ref{sec:dbms-setup}. This setup is expected to provide more practical insights into the computational cost reduction in real-world workloads than using synthetic functions. For comparing with proposed dynamic tree depth setting,  the maximum depth of the search tree is fixed at constant values of 1, 2, 3, and 4. It is important to note that the depth being 1 represents a special case where the search tree consists of a single root node, and the algorithm degrades to plain TuRBO because no partitions are split for guiding the local search. Performance data from previous experiments using dynamic adjustment are used, denoted by 'Adaptive (HiBO)'. 

In addition to performance measurements, considering the maximum tree depth is closely related to the tuning time cost, S-PITR is calculated to evaluate performance improvement relative to tuning time and safety factors for a more comprehensive evaluation. The results are presented in Figure \ref{fig:abs_depth}, from which we have the following findings:
\begin{itemize}
  \item Increasing tree depth from 1 to 3 in constant-depth search trees improves both performance and S-PITR, indicating the effectiveness of HiBO's partition-level guidance. The growing S-PITR shows that the additional computational cost of increasing tree depth is outweighed by the performance gains.

  \item Performance declines when the depth increases to 4, with results falling below those of plain TuRBO. This suggests that a tree depth of 4 introduces excessive bias toward limited space partitions, leading to over-exploitation. Additionally, the higher computational cost of constructing a 4-layer tree further reduces the S-PITR score. 

  \item Our proposed dynamic tree-depth adjustment achieved the highest performance improvements, with superior S-PITR scores. This demonstrates that our approach effectively balances exploitation and exploration with reasonable tuning time and failure rates. 
\end{itemize}

\begin{table}[t]
\vskip 0.15in
\begin{center}
\begin{small}

\bgroup
\def\arraystretch{1.16}%
\begin{tabular}{p{2.8cm}p{2.2cm}p{2.3cm}p{2.7cm}p{2.1cm}}
\toprule
& \textbf{Optim Exec (\%)} & \textbf{Config Load (\%)}    & \textbf{Workload Exec (\%)} &\textbf{S-PITR ($1e^{-3}$)} \\ 
\midrule
\textbf{Depth=1 (TuRBO)}    & 2.95 ± 1.07           & 29.16 ± 5.21           & 67.89 ± 4.87 &64.44 ± 6.80     \\ 
\textbf{Depth=2}  &3.98 ± 0.41           & 28.84 ± 3.40           & 67.18 ± 3.19    &67.68 ± 0.83         \\ 
\textbf{Depth=3}  & 4.21 ± 0.36           & 32.44 ± 1.22           & 63.35 ± 1.22   &69.19 ± 8.11          \\ 
\textbf{Depth=4}  &4.75 ± 1.04           & 32.24 ± 2.99           & 63.01 ± 3.01  &56.30 ± 8.64          \\ 
\textbf{Adaptive (HiBO)} & \textbf{4.31 ± 0.91 }          & 29.19 ± 3.28           & 66.50 ± 2.98    &\textbf{74.59 ± 3.64}        \\ 
\bottomrule
\end{tabular}
\egroup

\end{small}
\end{center}
\vskip -0.05in
\caption{Iteration time breakdown during applying HiBO for DBMS configuration tuning under different maximum tree depth settings. `Optim Exec` refers to the proportion of time spent on optimization algorithm execution. `Config Load` and `Workload Exec` similarly indicate the time proportion for loading the suggested configuration on the DBMS and executing the benchmark workload.}
\label{table:execution_times}
\vspace{-10pt}
\end{table}

Besides S-PITR, we provide a detailed breakdown of the iteration time on average during DBMS configuration tuning, which comprises the time for configuration loading, workload execution, and optimization execution. Table \ref{table:execution_times} summarizes the proportions for HiBO with different maximum tree depth settings. From the table, we observe that while HiBO introduces additional computational costs during the optimization phase compared to TuRBO, the increase remains minimal. Even with a fixed tree depth of 4, the added proportion of execution time is within 2\% on average per iteration. The adaptive tree depth control maintains computational costs between those of fixed depths of 3 and 4 while delivering the best performance improvements and highest S-PITR scores.

\section{Ablation Study: Temperature}\label{app: temp}

\begin{figure}[h]
    \centering
    \includegraphics[width=0.65\linewidth]{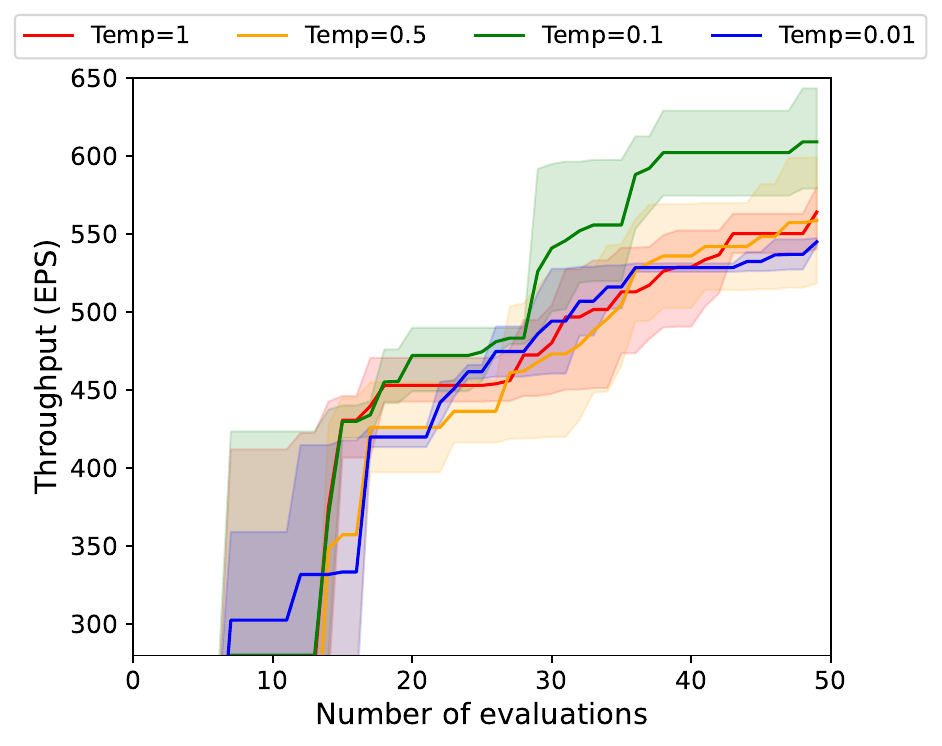}
    \caption{Performance (throughput) evaluation of PostgreSQL being tuned on SysBench by HiBO with different temperature settings.}
    \label{fig:dbms_temp}
\end{figure}

This section presents an ablation study on the effect of the hyperparameter $\tau$, which serves as the temperature in the Softmax (\ref{eq: softmax}). $\tau$ controls the influence of global-level space partitioning by adjusting the smoothness of the estimated distribution of sampling potential across partitions, where the two extreme cases of its value can lead to the following effects:
\begin{itemize}
    \item \textbf{Extremely high $\tau$}: Results in an overly smoothed distribution, reducing the impact of global-level partitioning and diminishing differences between partitions;
    \item  \textbf{Extremely low $\tau$}: Results in a peaky distribution and assigns overwhelming weight to the partition with the highest sampling potential, amplifying the navigator's influence.
\end{itemize}

We evaluated HiBO's performance on DBMS configuration tuning with $\tau$ values of 1, 0.5, 0.1, and 0.01. Other experiment settings follow those described in Section \ref{sec:dbms-setup} and Appendix \ref{app: expr_set}.

The evaluation results are presented in Figure \ref{fig:dbms_temp}. The results show that when temperature is set to 0.1, the performance improvement and improvement rate is evidently better than other temperature settings. While the other three temperatures have similar tuning effects with setting temperature to 0.01 resulting in slight degradation on final performance improvements. These findings suggest the existence of an optimal range for $\tau$. For tuning configurations of PostgreSQL towards workload SysBench, $\tau = 0.1$ strikes a good balance between these two ends of phenomenon and achieved better DBMS performance improvements on SysBench.

\section{Evaluation on Additional Real-world Benchmarks} \label{app: eval_real}
\begin{figure}[t]
    \centering
    \includegraphics[width=0.96\linewidth]{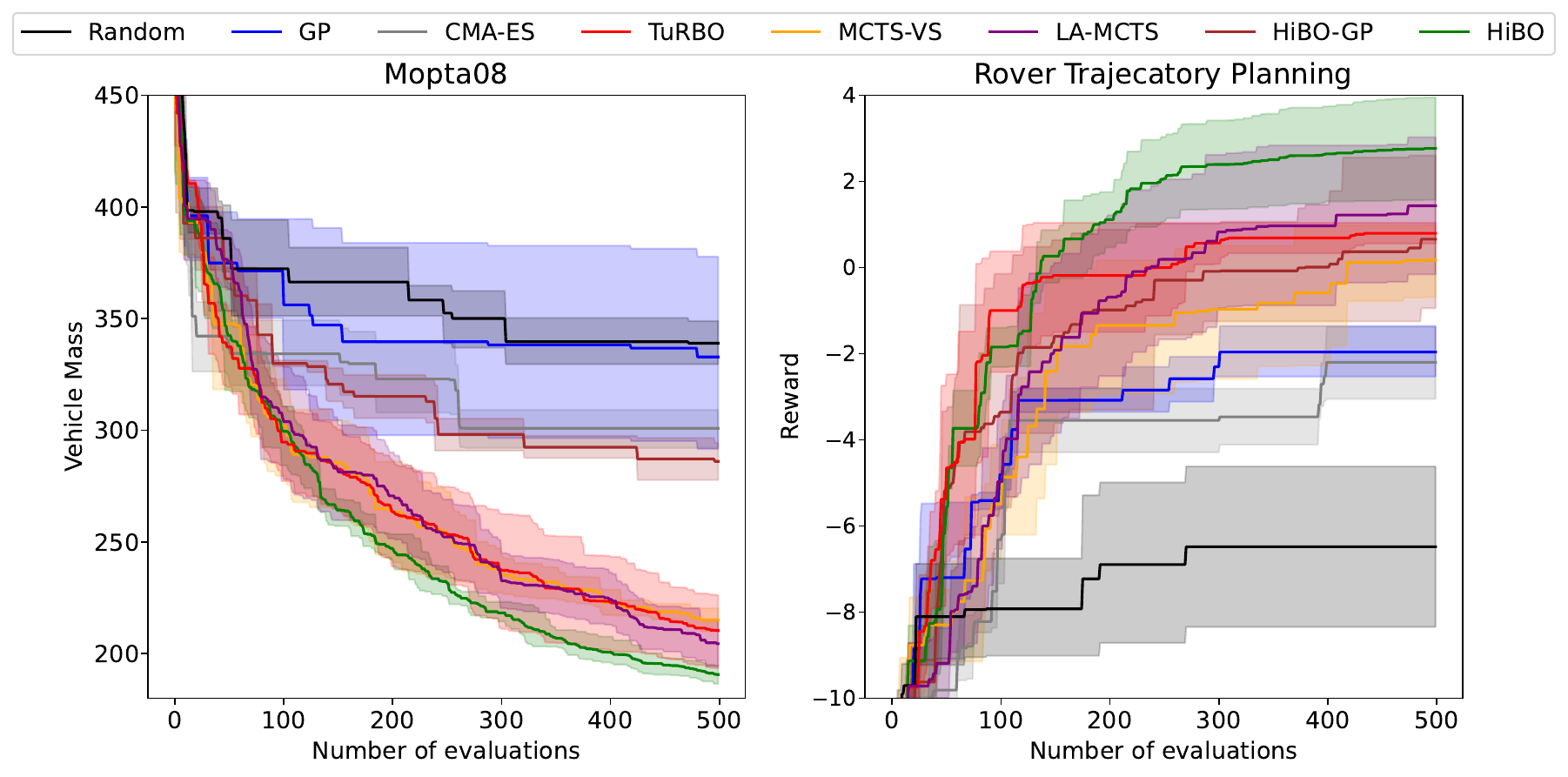}
    \caption{Evaluation results of algorithms on Mopta08 (Left) and Rover Trajectory Planning (Right).}
    \label{fig:real_world}
\end{figure}

Though DBMS configuration tuning has been introduced as a unique challenging benchmark for evaluating BO-based algorithms on high-dimensional search space, we still evaluated HiBO on two widely recognized real-world benchmarks in prior works \citep{eriksson2019turbo, eriksson2021saasbo, shen2023vsbo, rashidi2024CTS} to further validate its performance, which include the following two benchmarks:
\begin{itemize}
    \item  Mopta08 (124D; \citet{jones2008mopta}): This task involves minimizing the mass of a vehicle by optimizing 124 design variables that control materials, gauges, and vehicle shape. The optimization is subject to 68 constraints, including performance, safety, and feasibility requirements. Mopta08 is a high-dimensional, constrained optimization problem widely used to benchmark BO methods in real-world-inspired engineering design scenarios.
    \item Rover Trajectory Planning (60D; \citet{wang2017rover}): In this task, the objective is to maximize the total reward collected by a rover along a planned trajectory. The trajectory is defined by optimizing the coordinates of 30 points in a 2D plane, resulting in a 60-dimensional search space. This benchmark reflects practical challenges in robotics, particularly in path planning and navigation under constraints.
\end{itemize}

The results for these additional benchmarks are presented in Figure \ref{fig:real_world}. As shown in Figure 8, HiBO demonstrates competitive performance across both tasks, achieving an 20 more units of vehicle mass on Mopta08 and an 1 more reward value on Rover Trajectory Planning than LA-MCTS on average. Furthermore, HiBO-GP consistently outperforms standard GP-BO in these benchmarks. Specifically, HiBO-GP achieves over 25\% less vehicle mass on Mopta08 and approximately 2 additional reward values on Rover Trajectory Planning compared to vanilla GP-BO. These results further underscore HiBO’s effectiveness and its ability to tackle complex, high-dimensional optimization problems frequently encountered in practical applications.

\section{Limitations}\label{app: discussion}
We summarized the following limitations with HiBO based on reasoning and experiment results:
\begin{itemize}
    \item \textbf{Performance Improvment on Dense Benchmarks}. One limitation of HiBO lies in its relatively less pronounced improvement over the local optimizer on dense high-dimensional benchmarks compared to sparse benchmarks. While HiBO demonstrates significant performance gains on sparse benchmarks, its improvement is less evident on dense tasks where all dimensions are uniformly effective such as Levy-200D and Rastrigin-200D). Our interpretation is that in sparse spaces, HiBO efficiently identifies promising regions to guide the local optimizer and it is one of its greatest advantages. While in dense spaces, it requires more samples to achieve accurate estimations across the entire space than in sparse scenarios. But HiBO still converges efficiently towards the optimal point in these tasks and achieves the best in Levy-200D and Rastrigin-200D. 
    \item \textbf{Unobservable Configurations}. HiBO may also struggle in scenarios where certain critical configuration variables are hidden or unobservable during optimization. Such situations can arise in practical tasks where parts of the configuration space are inaccessible due to hardware constraints, legacy systems, or incomplete problem specifications. In these cases, the global-level navigator may build an inaccurate model of the search space, leading to suboptimal partitioning and sampling. 
\end{itemize}

\end{document}